\documentclass[10pt,twocolumn,letterpaper]{article}

\usepackage[pagenumbers]{cvpr} 

\usepackage[dvipsnames]{xcolor}

\definecolor{cvprblue}{rgb}{0.21,0.49,0.74}
\usepackage[pagebackref,breaklinks,colorlinks,citecolor=cvprblue]{hyperref}

\newcommand{\x}{\mathbf{x}}
\newcommand{\y}{\mathbf{y}}

\usepackage{bbm}
\usepackage[ruled]{algorithm2e}
\usepackage{multirow}
\usepackage{tabularray}

\SetKwInput{KwInput}{Input}
\SetKwInput{KwOutput}{Output}

\renewcommand{\paragraph}[1]{\vspace{0.5em}\noindent\textbf{#1}}

\def\paperID{14852}
\def\confName{CVPR}
\def\confYear{2024}

\title{Multi-Modal Hallucination Control by Visual Information Grounding}

\author{
Alessandro Favero\thanks{Work done during an internship at AWS AI Labs.}
\hspace{1.5em}
Luca Zancato
\hspace{1.5em}
Matthew Trager
\hspace{1.5em}
Siddharth Choudhary\\
Pramuditha Perera
\hspace{1.5em}
Alessandro Achille
\hspace{1.5em}
Ashwin Swaminathan
\hspace{1.5em}
Stefano Soatto \vspace{0.5em}
\\
AWS AI Labs \vspace{0.2em}\\
{\tt\small alessandro.favero@epfl.ch}\\
{\tt\small \{zancato,mttrager,sidchoud,pramudi,aachille,swashwin,soattos\}@amazon.com}\\
}

\begin{document}
\maketitle

\begin{abstract}
\noindent
Generative Vision-Language Models (VLMs) are prone to generate plausible-sounding textual answers that, however, are not always grounded in the input image. We investigate this phenomenon, usually referred to as ``hallucination'' and show that it stems from an excessive reliance on the language prior. In particular, we show that as more tokens are generated, the reliance on the visual prompt decreases, and this behavior strongly correlates with the emergence of hallucinations. To reduce hallucinations, we introduce Multi-Modal Mutual-Information Decoding (M3ID), a new sampling method for \textit{prompt amplification}. 
M3ID amplifies the influence of the reference image over the language prior, hence favoring the generation of tokens with higher mutual information with the visual prompt. M3ID can be applied to any pre-trained autoregressive VLM at inference time without necessitating further training and with minimal computational overhead. If training is an option, we show that M3ID can be paired with Direct Preference Optimization (DPO) to improve the model's reliance on the prompt image without requiring any labels. Our empirical findings show that our algorithms maintain the fluency and linguistic capabilities of pre-trained VLMs while reducing hallucinations by mitigating visually ungrounded answers. Specifically, for the LLaVA 13B model, M3ID and M3ID+DPO reduce the percentage of hallucinated objects in captioning tasks by 25\% and 28\%, respectively, and improve the accuracy on VQA benchmarks such as POPE by 21\% and 24\%.
\end{abstract}

\vspace{-12pt}

\section{Introduction}

\begin{figure}[h!]
    \includegraphics[width=\linewidth]{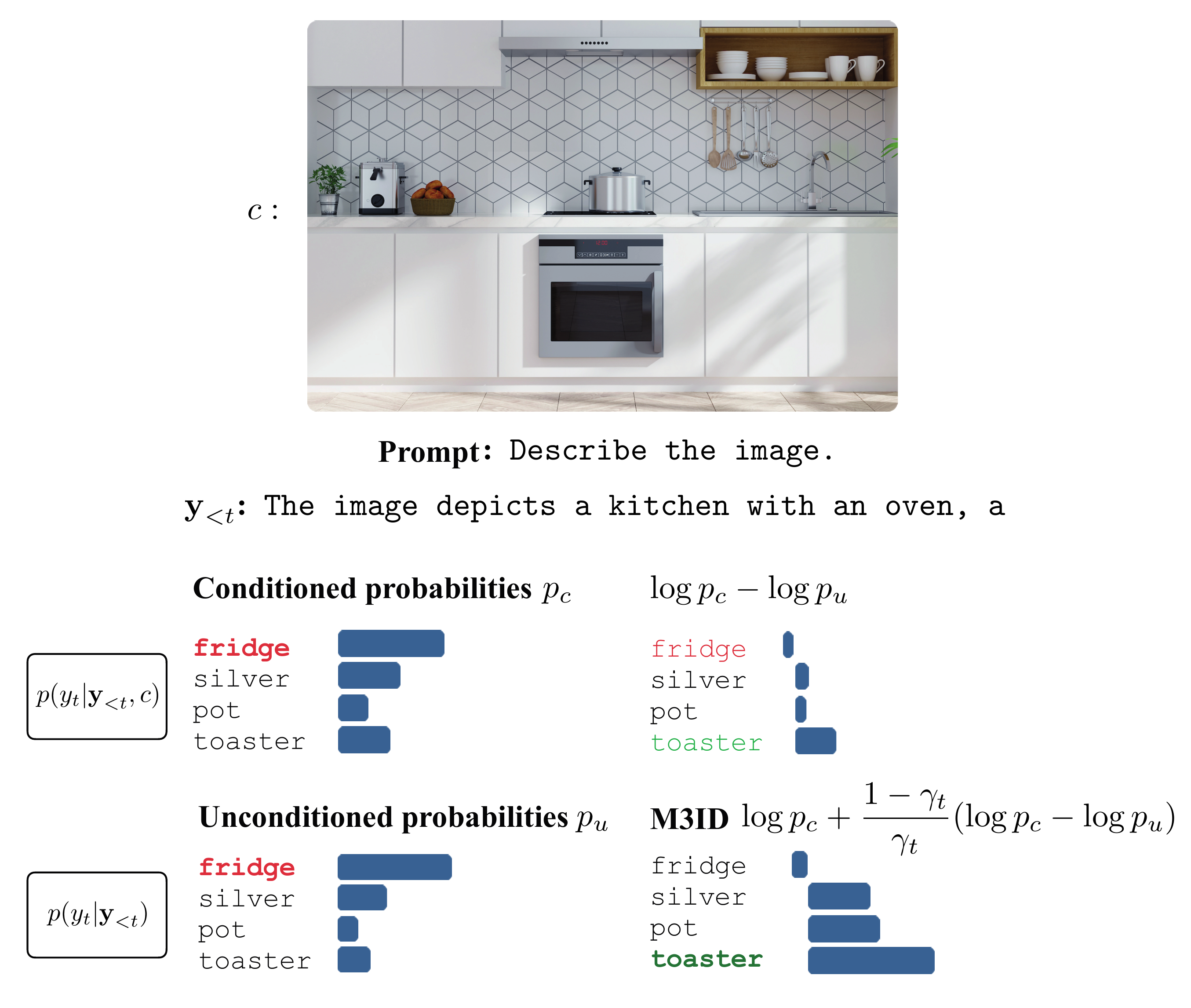}
    \caption{\textbf{Multi-Modal Mutual Information Decoding (M3ID).} 
    Given a VLM $p$, an image $c$, and a text prompt, M3ID intervenes in the generative distribution by finding tokens that ``surprise'' the unconditioned VLM (\ie, the VLM without the image prompt). M3ID amplifies the conditioned directions that are not already predicted by the unconditioned model more as new tokens are generated by leveraging a progressively smaller $\gamma_t$. In the example, the VLM assigns a high likelihood to the hallucinated object \textit{fridge}, over-relying on its unconditioned language prior. Instead, M3ID assigns a high likelihood to \textit{toaster}, which is present in the image.
    }
    \label{fig:splash}
    \vspace{-0.5cm}
\end{figure}

Recent autoregressive Vision-Language Models (VLMs) have shown remarkable multimodal capabilities \cite{llava, dai2023instructblip}. However, VLMs, similarly to large language models (LLMs), are prone to ``hallucinations'' -- generating plausible-sounding answers without factual basis, leading to potentially \textit{ungrounded} or fabricated information \cite{MMI_decoding_chatbots_15}. Consequently, the community has been developing ever so complex and expensive alignment algorithms involving direct human supervision \cite{RLHFouyang2022training} and often brittle prompt engineering methods (``model begging'') \cite{chain_of_thoght_wei2022chain}.

\looseness=-1 In this work, we propose to investigate hallucinations in VLMs through a quantifiable measure of visual \textit{prompt dependency}. We assess whether a model output is ungrounded with respect to a visual prompt by comparing its likelihood with the likelihood of generating the same output without the visual information. In general, our measure of visual prompt dependency quantifies the extent to which a token is generic or specific to a visual signal. Importantly, low visual prompt dependency does not necessarily imply that a token is hallucinated, as it may occur in linguistically essential elements like conjunctions and prepositions. Yet, unlike the notion of hallucinations, which requires human judgment, our prompt dependency measure is always well-defined even without any ground truth annotations.

Our first contribution is to empirically demonstrate that the visual prompt dependency measure decreases as more tokens are generated. In \cref{fig:fading-memory}, we show that, as more tokens are generated, the conditioning information gets diluted and ``forgotten'' or ignored by the model, possibly leading to more hallucinations. 
We refer to this effect as \textit{conditioning dilution} or \textit{fading memory effect}.

To counteract this, we propose an intervention on the generative distribution of a VLM that maximizes visual prompt dependency at inference time. We show that our method for \textit{prompt amplification} maximizes the mutual information between the text output tokens and the visual prompt, effectively rescaling the image-conditioned component against the unconditioned distribution. We name our inference-time intervention on the generative distribution \textit{Multi-Modal Mutual Information Decoding} (M3ID). M3ID is applicable to any off-the-shelf model without additional training or access to model weights, offering a low computational overhead alternative to standard decoding algorithms \cite{llama, llava}. Our results show that M3ID enhances the dependence on the visual prompt and reduces the number of hallucinations across various benchmarks while preserving the linguistic fluency of the original model.

\looseness=-1 Additionally, for users with access to model weights, we propose a training objective to further ground the model outputs on visual prompts. This approach involves recasting our prompt amplification objective as a preference optimization problem, using Direct Preference Optimization (DPO) \cite{rafailov2023direct} to align the pre-trained VLM. This method aims to increase the likelihood of continuations with a higher visual prompt dependency measure and allows to learn a better generation policy, further reducing the conditioning dilution effect and hallucinations.

In summary, our main contributions are as follows: 
\begin{enumerate}
    \item We propose a visual prompt dependency measure (PDM) to assess whether a model output is ungrounded with respect to the visual input. Empirically, we demonstrate that PDM decreases as more tokens are generated making the generations more likely to be hallucinated.
    \item \looseness=-1 We introduce M3ID, a training-free intervention on the generative distribution of autoregressive VLMs which improves visual grounding and reduces hallucinations by amplifying the importance of the visual prompt over the language prior. In addition, we extend M3ID with DPO for multi-modal preference optimization to further improve visual grounding.
    \item We show that applying M3ID or DPO reduces the percentage of hallucinated objects in \textit{captioning tasks} \cite{rohrbach2018object} by 25\% and by 28\%, respectively and improves accuracy on the POPE \cite{li2023evaluating} VQA hallucination benchmark by 21\% and 24\% over the base model.
\end{enumerate}

\section{Related work}

\paragraph{Hallucinations in VLMs.} 
\looseness=-1 A recent line of work introduced VLMs obtained by \textit{grafting} visual encoders into pre-trained Large Language Models (LLMs) \cite{llava, dai2023instructblip, zhu2023minigpt}. Forcing a pre-trained LLM to learn to ``see'' by reading vision tokens from a pre-trained vision backbone has proven successful and the resulting models show remarkable vision-language understanding capabilities on many multi-modal tasks \cite{llava, sun2023aligning}. However, while inheriting strong linguistic capabilities and fluency from their base LLM, grafted VLMs also inherit the tendency to produce ungrounded or fabricated information \cite{huang2021factual, bang2023multitask}. This is commonly referred to as ``hallucination'' in the recent machine learning literature \cite{PMI_for_summarization, huang2021factual, bang2023multitask}. In particular, several recent works empirically reported a sharp tendency of grafted VLMs to report objects not grounded on the visual information when probed with questions on the image content \cite{li2023evaluating, wang2023evaluation, bang2023multitask, huang2021factual}. The authors of \cite{li2023evaluating} observed that this phenomenon is especially pronounced with objects that are either common or frequently appear together in the datasets used at training time. Furthermore, the authors of \cite{liu2023aligning} suggest that VLMs often fail to correctly follow instructions involving absent objects and propose a new instruction-following training objective to enhance model alignment and robustness in the face of uncertainty. Parallel to our work, \cite{zhou2023analyzing} delves into the factors contributing to object hallucinations in VLMs, exploring aspects like object co-occurrence, model uncertainty, and spatial position of hallucinations in the sentence. They propose a post-hoc algorithm to identify and correct hallucinations in VLM-generated content.

\paragraph{Context-dependent decodings.} 
Decoding algorithms can be classified as either search or sampling algorithms \cite{contrastive_decoding_li23}. Current search methods (\eg, greedy and beam search) can produce factually accurate generations but usually suffer from tedious and repetitive continuations. Sampling methods, on the other hand, (like nucleus \cite{holtzman2019nucleous_sampling}, or typical decoding \cite{meister2023typical}) produce more diverse text, but suffer from topic drift. To solve these shortcomings, various decoding methods have been proposed to enhance reasoning and factual accuracy in generative autoregressive language models. Approaches such as context-aware decoding \cite{shi2023trusting} and Pointwise Mutual Information (PMI) decoding \cite{PMI_for_summarization, nandwani2023pointwise} focus on aligning the outputs of language models with the intended context or factual data. These strategies aim to reduce hallucinations in standard natural language processing tasks such as summarization. Mutual-information-based decoding techniques have proven to be helpful in steering the generation of LLMs to stay faithful to the given input text \cite{PMI_for_summarization}, or to promote diversity or relevance in neural dialogue models \cite{MMI_decoding_chatbots_15}. Our work is the first to use mutual information to improve multi-modal grounding and to vary the penalization of marginally likely outputs in a time-varying fashion to counteract the progressive forgetting of the visual prompt. 

\begin{figure}
    \centering
    \includegraphics[width=1.\linewidth]{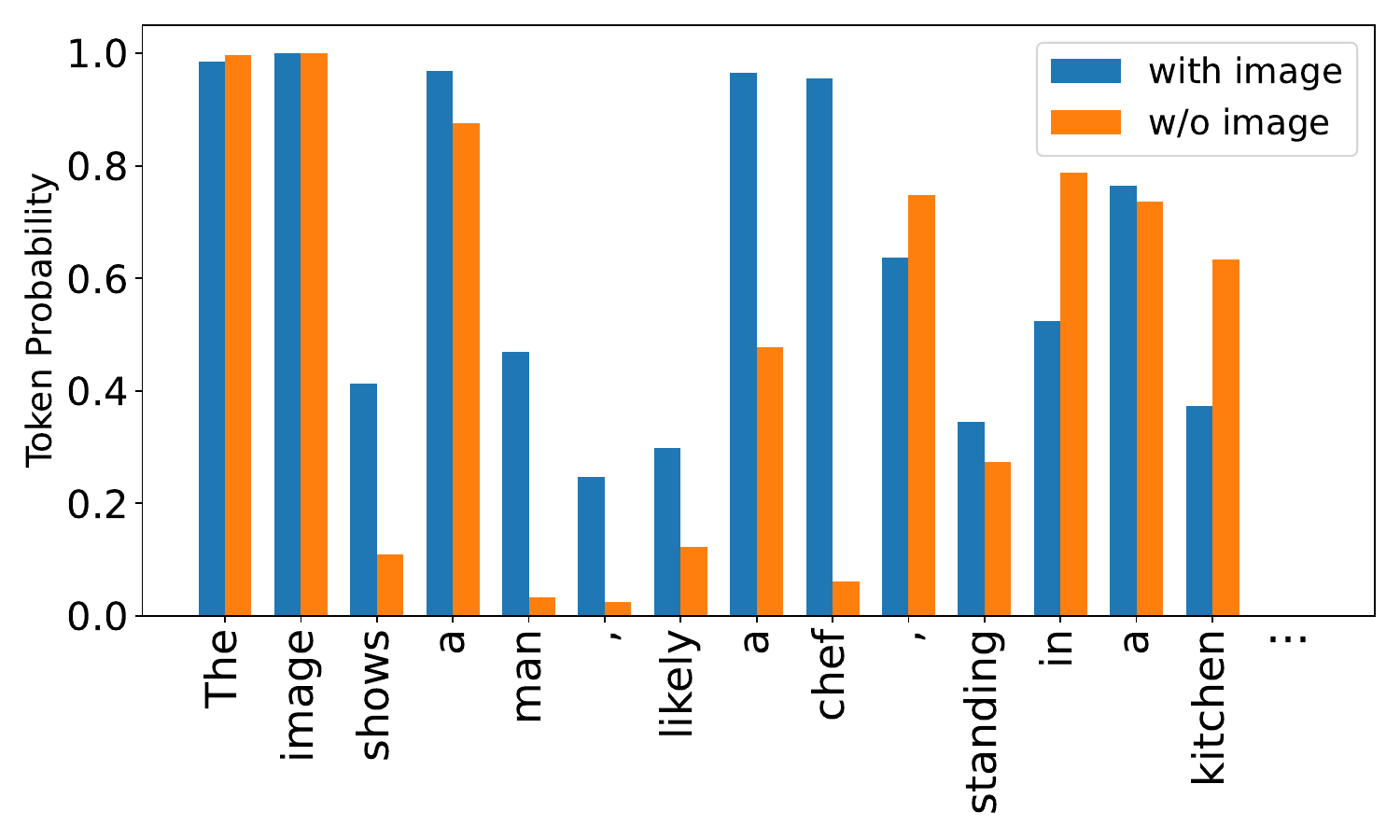}
    \vspace{-0.75cm}
    \caption{\textbf{How much does the conditioning prompt ``surprise'' the unconditioned predictor?} We report the likelihood that the conditioned and unconditioned models assign to each token in the string ``The image ...''. The probability gap increases on tokens for which visual information is required (\eg objects and attributes), while it decreases for punctuation and articles. Note that both models mostly agree when sufficient visual information has already been extracted and is present in the caption (\eg ``kitchen'' is very likely for the unconditioned model as well). 
    See \cref{sec:PDM-appendix} for a plot with our prompt dependency measure. 
    }
    \vspace{-0.3cm}
    \label{fig:rank_distance}
\end{figure}

\section{Analysis of hallucinations in VLMs} 
\label{sec:VLMs_are_fading_memory_sys_examples}

\looseness=-1 We start by investigating hallucinations in VLMs. We introduce a visual \textit{prompt dependency measure} (PDM), which will later motivate our algorithms (\cref{sec:methods}). The main aim of this section is to assess whether a model output, such as a token, is hallucinated or not by comparing its likelihood with the likelihood of that same output being generated without the relevant conditioning factor, in this case, the image.

We consider decoding approaches for open-ended language generation of VLMs, where the model receives an input image and an input textual prompt and aims to generate a fluent and coherent continuation conditioned on both the image and the text. We denote probabilistic generative VLMs over a vocabulary of text tokens $\mathcal{V}$ with $p$ and denote the probability of a token $y \in \mathcal{V}$ given a textual prompt $\x$ and an image context $c$ as $p(y|\x, c)$. The goal of the VLM is to leverage the information contained in the input image $c$ to provide a continuation $\y=[y_0, \dots, y_T]$ of the input prompt $\x$, \eg, ``Describe this image in detail.''. The probability of a valid sequence $\y$ can be computed as $p(\y|\x, c) = \prod_{t=1}^T p(y_t | \y_{<t}, \x, c)$, where $\y_{<t} \triangleq [y_0, \dots, y_{t-1}]$. 

\begin{figure}
\begin{minipage}{.5\linewidth}
  \centering
    \includegraphics[width=\linewidth]{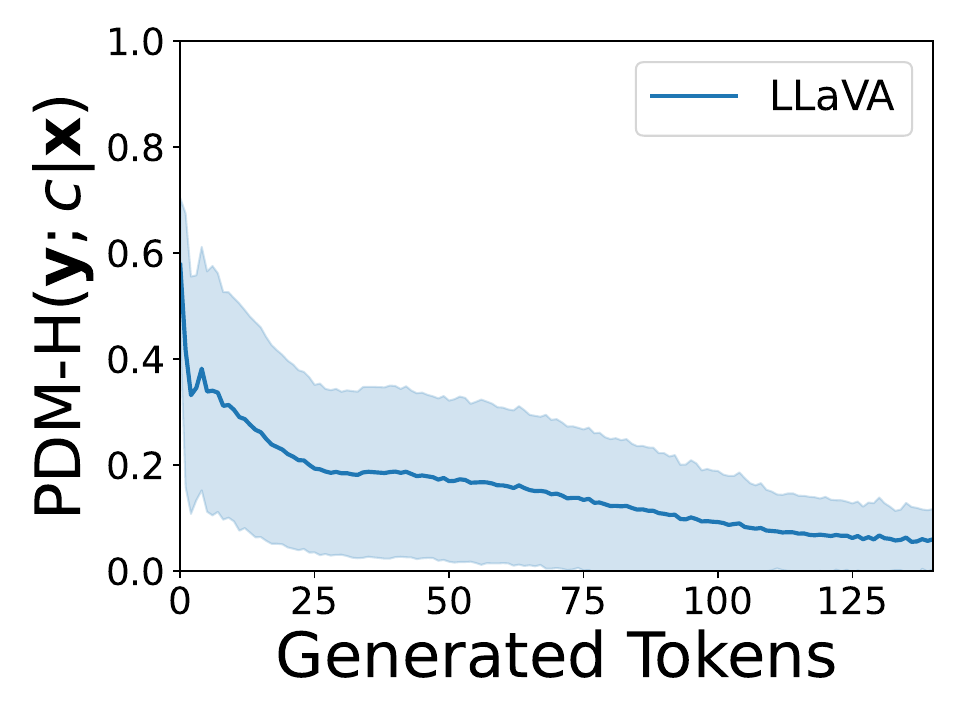}
\end{minipage}%
\begin{minipage}{.5\linewidth}
    \centering
    \includegraphics[width=\linewidth]{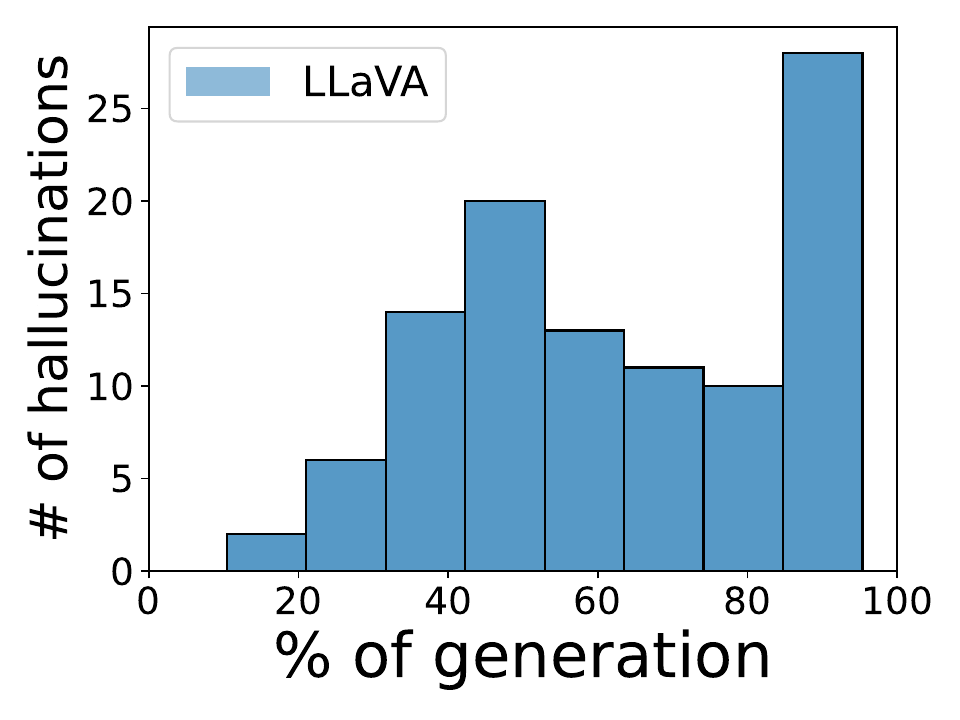}
\end{minipage}
    \caption{
    \textbf{The influence of the image conditioning decreases as we generate more tokens.}
    \textbf{(Left) Conditioning dilution.} 
    We report the average of the prompt dependency measure (PDM-H) and its standard deviation for different synthetic captions generated by LLaVA on MS COCO's validation split.
    We see that the influence of the image over the next token prediction decreases as we generate more.
    This suggests that the information within VLM's visual prompt gets diluted and fades away as more tokens are generated.
    \textbf{(Right) Frequency of hallucinated objects as a function of the token position.} 
    We report the number of non-existent objects present on the same synthetic captions as a function of the number of generated tokens. Note that very few objects are hallucinated for tokens near the visual prompt, while their number increases as more tokens are generated and with a smaller PDM.
    }
    \vspace{-0.3cm}
    \label{fig:fading-memory}
\end{figure}

We propose to study hallucinations on VLMs using PDMs defined as follows
\begin{equation}
    \text{PDM}(\y_{<t} ; c | \x) \triangleq {\rm dist}\Big(p(\cdot| \y_{<t}, \x, c),\, p(\cdot| \y_{<t}, \x) \Big) \nonumber
\end{equation}
where ${\rm dist}$ is any distance measure between probability distributions, such as Hellinger, total variation, or KL. PDMs quantify how generic or context-specific a language model's output is. In particular, a high $\text{PDM}(\y_{<t} ; c | \x)$ indicates that the token $y_t$ is strongly associated with a specific input prompt, while a low PDM suggests that the token is more prompt-neutral or prompt-agnostic. Depending on the choice of the distance function, PDMs highlight different aspects of the generative distribution. We will mainly use $\text{PDM-H}$ based on the Hellinger distance defined as $H(p,q)=2^{-1/2} \sqrt{\sum_{i=1}^k (\sqrt{p_i}-\sqrt{q_i})^2}$, where $p=(p_i)_{i\in[k]}$ and $q=(q_i)_{i\in[k]}$ are discrete probability distributions. We refer the reader to the Supplementary Material for an analysis of the impact of other distances.

\paragraph{Contextual pressure.}\label{sec:contextual_pressure}
\cref{fig:rank_distance} shows the conditional and unconditional prediction likelihoods on a single caption. First, we note that tokens that are required for linguistic fluency, like prepositions and conjunctions, are highly predictable by both the conditioned and the unconditioned models. This pattern also emerges when the model generates tokens to compose ``fine-grained'' objects. For example, when the model generates ``Peanut butter'', both the VLM and the LLM can correctly predict the last token even without looking at the image and only having access to a sufficiently long truncation (\eg ``Peanut bu''). Therefore, in these circumstances, $p(y_t|\y_{<t}, \x, c)$ and $p(y_t|\y_{<t}, \x)$ can be very close irrespective of the number of generated tokens and without implying a higher hallucination risk. We will refer to this phenomenon as \textit{contextual pressure}. 

\looseness=-1 \paragraph{Conditioning dilution and hallucinations.}\label{sec:conditioning_dilution}
In \cref{fig:fading-memory} we show that PDM-H decreases as more tokens are generated, indicating that the visual information gets diluted and neglected by the model throughout the generation process. This phenomenon suggests that the conditioned model distribution gets closer to the unconditioned one, the language prior, as we generate more tokens.
In other words, the VLM places high probability mass on marginally likely tokens that are mostly explained by their language priors, $p(y_t|\y_{<t}, \x, c) \to p(y_t|\y_{<t}, \x)$ as $t$ increases. 
We refer to this phenomenon as \textit{conditioning dilution} or \textit{fading memory effect}. While conditioning dilution is not necessarily an issue (for instance, if the generated text has already extracted all of the relevant information from the context), one ideally would want to prevent the model from only relying on language priors based on the generated caption which may lack important details and facilitate hallucinations. Notice that whereas PDMs are useful for measuring the importance of the visual prompt, their values do not fully characterize the likelihood of generating ungrounded tokens. Therefore, in \cref{fig:fading-memory} we empirically test how much our prompt dependency measure correlates with the number of hallucinated objects generated by a SOTA VLM (LLaVA \cite{llava}) on the MS COCO dataset \cite{coco2014}. 
In the figure, we report the number of non-existent objects on captions generated by LLaVA as a function of the number of generated tokens. Note that very few objects are hallucinated near the input context, while their number increases as the PDM gets smaller. This motivates us to intervene in the generative distribution to maximize the PDM in order to reduce hallucinations.

\section{Methods}\label{sec:methods}

Building on insights from \cref{sec:VLMs_are_fading_memory_sys_examples}, we formalize how to model the generation of VLMs as a fading memory process and how to intervene in the generation to prevent the VLM from ``forgetting'' the visual prompt $c$. In this section, we denote the log probabilities as $l(y_t|\y_{<t}, \x, c) \triangleq \log p(y_t|\y_{<t}, \x, c)$. 

\subsection{M3ID: Improving grounding at inference time}

As we have shown in \cref{sec:conditioning_dilution}, within their prediction horizon, VLMs can be modeled as fading memory autoregressive systems \cite{fading_memory_systems, fading_memory_sys_id}: the more they generate the smaller the impact of the visual information on the generated tokens. As such, the longer the generation, the higher the reliance on the language priors which favor continuations not specifically grounded on the conditioning signal.

\paragraph{Preventing conditioning dilution.}
Under the fading memory assumption, we model the conditioned log probabilities $l(y_t | \y_{<t}, \x, c)$ (for example, LLaVA \cite{llava}) as an interpolation between the unconditioned model and a model $l^*$ that does not ``forget'' old context as more tokens are generated. 
\begin{equation}\label{eq:fading_memory_model}
    l(y_t | \y_{<t}, \x, c) = \gamma_t \, l^*(y_t | \y_{<t}, \x, c) + (1-\gamma_t) \, l(y_t|\y_{<t}, \x)
\end{equation}
where $\gamma_t \in [0, 1]$ is a mixing coefficient which monotonically decreases over time. When $\gamma_t$ is small, the conditional distribution is mostly explained without providing the input image, and the conditioned distribution ``forgets'' it, while for higher $\gamma_t$, the input image becomes more relevant. We take $\gamma_t \triangleq \exp(-\lambda t)$, $\lambda$ models the rate of change of $\gamma_t$ and defines the fading memory rate (\textit{forgetting rate} \cite{fading_memory_systems, fading_memory_sys_id}).

Given observations from the conditioned distribution $l_c \triangleq l(y_t | \y_{<t}, \x, c)$ and the unconditioned one $l_u \triangleq l(y_t | \y_{<t}, \x)$, our goal is to find an estimate $\hat{l}^*$ of the latent generative distribution $l^*$ which does not forget the past.

To do so, we assume that $l^*$ is a perturbation of the conditioned distribution, $l^* = l_c + \Delta$, where $\Delta$ can be assumed to be a random variable with zero mean and bounded variance. Therefore we rewrite our model in \cref{eq:fading_memory_model} as: 
\begin{equation}\label{eq:noisy_measurement_function}
    (1-\gamma_t) (l_c - l_u) = \gamma_t \Delta.
\end{equation}

\looseness=-1 Over the time index $t$, \cref{eq:noisy_measurement_function} is a stochastic process across tokens whose variance decreases over time. 
Hence, we can estimate the optimal intervention on $l_c$ to counteract the fading memory effect and get closer to $l^*$ as $\hat{l}^* = l_c + \hat{\Delta}$.
We use measurements from \cref{eq:noisy_measurement_function} to estimate the correction term $\hat \Delta$. This brings us to the optimal intervention: 
\begin{equation}\label{eqn:optimal_intervention}
    \hat{l}^* = l_c + \frac{1-\gamma_t}{\gamma_t} (l_c - l_u).
\end{equation}
Note that the optimal sampling distribution $l^*$ is approximately proportional to $l_c$ at the beginning of the generation (\ie, when $\gamma_t$ is close to 1), which translates to sampling from the conditioned distribution alone. On the other hand, far from the input prompt (\ie, $\gamma_t \to 0$) the optimal sampling distribution becomes proportional to $l_c - l_u$. This amplifies tokens proposed by $l_c$ that ``surprise'' the unconditional policy $l_u$ and can be thought of as a way to ameliorate the conditioning dilution phenomenon described in \cref{sec:conditioning_dilution}. 
From an information theory viewpoint, the second scenario corresponds to maximizing the pairwise mutual information between the visual input and the text tokens instead of maximizing the log-likelihood of the text tokens alone. In fact, $\max_\y \log \frac{p(c,\y|\x)}{p(c)p(\y|\x)}=\max_\y \log\frac{p(\y|\x,c)}{p(\y|\x)}=\max_\y l_c - l_u$ \cite{PMI_for_summarization, MMI_decoding_chatbots_15}.

\paragraph{Accommodating for contextual pressure.} 
Notice, however, that the contextual pressure forces $l_c$ and $l_u$ to be similar irrespective of the number of generated tokens. As such, penalizing the language prior $l_u$ by amplifying $l_c - l_u$ indiscriminately would sometimes also penalize correct but obvious tokens that do not require the input image to be inferred (like prepositions or conjunctions). To avoid such a scenario, we suppress our intervention in \cref{eqn:optimal_intervention} when the contextual pressure is high. Specifically, if the conditioned model $l_c$ is highly \textit{confident} on the next token (\ie, it has the maximum probability above a threshold $\alpha$ \cite{contrastive_decoding_li23}) we do not apply the correction term $l_c - l_u$. 

\paragraph{Multi-Modal Mutual Information Decoding (M3ID).}
Putting everything together, our Multi-Modal Mutual Information Decoding \cref{algo:m3id} for generation is given by:
\begin{equation}\label{eq:m3id_equation}
  y_{t} = \arg \max_{y \in \mathcal{V}} \hat l^*(y | \y_{<t}, \x, c)
\end{equation}
where  $\hat l^* = l_c + \mathbbm{1} \Big[\max_k (l_c)_k < \log \alpha \Big] \frac{1 - \gamma_t}{\gamma_t} (l_c - l_u)$. M3ID can be applied to different search algorithms like greedy search (as in \cref{eq:m3id_equation}) or beam search. 

\subsection{M3ID+DPO to learn more grounded policies}

With access to compute and model weights, we can optimize the model to output continuations that are more grounded to the image content. In this section, we reframe this goal into a \textit{preference optimization problem}, where the objective is to prefer grounded continuations over ungrounded ones and fine-tune the VLM using this objective. 

\paragraph{Multi-modal preference optimization.} 
Consider two different continuations $\y_w$ and $\y_l$ of the same prompt $\x$ for the image $c$, with $\y_w$ being preferred over $\y_l$. 
Direct Preference Optimization (DPO) \cite{rafailov2023direct} is an alignment technique that aims at learning a generative policy that is more likely to generate continuations \textit{similar} to $\y_w$ rather than $\y_l$.
In particular, given a dataset $\mathcal{D}$ containing preference data pairs $(\y_w, \y_l)$, DPO minimizes the following loss function, 
\begin{align}\label{eq:dpo}
    \mathcal{L}_{\rm DPO} = - \mathbb{E}_{(c,\x,\y_w,\y_l)\sim \mathcal{D}} &\left[ \log \sigma \left( \beta \log \frac{p_\theta(\y_w|c,x)}{p_{\rm ref}(\y_w|c,x)} \right. \right. \nonumber \\
     &- \left. \left. \beta \log \frac{p_\theta(\y_l|c,x)}{p_{\rm ref}(\y_l|c,x)} \right)\right], \nonumber
\end{align}
where $p_{\rm ref}$ denotes the VLM before DPO training and $p_\theta$ the model being trained. 
In short, the intuition behind this objective is to increase the likelihood of the completion $\y_w$ with respect to the base model $p_{\rm ref}$ while decreasing the likelihood of generating $\y_l$ relative to the base model.

\looseness=-1 \paragraph{Generating multi-modal preference data}
The success of DPO hinges on the quality of the pairs in the dataset $\mathcal{D}$.  Therefore, we propose to fine-tune a pre-trained VLM using the DPO objective while ensuring that the preferred continuations are more grounded to the visual information. To do so, we generate preferred continuations by sampling from the pre-trained conditioned distribution improved with M3ID $\y_w \sim p^*(\y | \x, c)$, while we generate negative continuations by sampling from the unconditioned distribution $\y_l \sim p(\y | \x)$. However, notice that generating text continuations from $\x$ (\eg, ``Describe the image.'') with the unconditioned model leads to negative continuations $\y_l$ that are often very different from the image caption and thus provide a little signal in the DPO objective. To overcome this limitation we append to $\x$ the first sentence generated by the conditioned VLM to restrict the set of possible continuations. 

{
\begin{algorithm}[t]
\KwInput{VLM $p$, textual prompt $\x$, image $c$, threshold on confidence $\alpha$, forgetting rate $\lambda$}
\KwOutput{Generated string $\y$ conditioned on $\x$ and $c$}
$y_0 = \text{BOS}$, $t = 1$

\While{$y_{t} \neq {\rm EOS}$}{
 $\gamma_t \gets \exp(-\lambda t)$
 
 $l_c \gets \log p(y | \y_{<t}, \x, c)$
 
 $l_u \gets \log p(y | \y_{<t}, \x)$

 $\hat l^* = l_c + \mathbbm{1} \Big[\max_k (l_c)_k < \log \alpha \Big] \frac{1 - \gamma_t}{\gamma_t} (l_c - l_u)$

  $y_{t} = \arg \max_{y \in \mathcal{V}} \hat l^*(y | \y_{<t}, \x, c)$
}
\caption{M3ID}
\label{algo:m3id}
\end{algorithm}
\vspace{-0.3cm}
}

\section{Experiments}\label{sec:experiments}

We evaluate our methods on standard captioning and VQA benchmarks on MS COCO \cite{coco2014}, an object detection and captioning dataset with annotations for 80 object categories. We use ground-truth annotations to measure the number of ungrounded objects that are predicted by our models. 

\paragraph{Architecture.} 
Our method applies to any VLM, as long as it is possible to drop the visual conditioning to compute the language-only prior. In the following, we shall mainly use the LLaVA architecture \cite{llava}. LLaVA is an open-source VLM connecting a frozen pre-trained open-set visual encoder with a pre-trained large language model via a trainable linear mapping layer.  As a vision encoder, we use CLIP ViT L-14 \cite{clip}. As a large language model, we use the 7B and 13B versions of Vicuna v1.3 \cite{vicuna}, which are instruction-tuned from LLaMA \cite{llama}. All our base models have been trained on LCS-558K \cite{liu2023improved} and LLaVA-Instruct-80K \cite{llava}. We follow the default query format used in LLaVA-v1.3 for the input data \cite{llava}. 
Furthermore, in \cref{sec:blip} we extend our evaluations to InstructBLIP \cite{dai2023instructblip}, which introduces a Q-Former to bridge the vision and language modalities.

\paragraph{Baselines.}
For our baseline methods, we consider generation using standard multinomial sampling \cite{llama, llava} where the next token is randomly sampled from the prediction distribution with a given temperature parameter (0.2 for LLaVA \cite{llava}). Furthermore, we compare M3ID with other training-free methods for autoregressive models like PMI \cite{PMI_for_summarization} and Contrastive Decoding \cite{contrastive_decoding_li23}.  PMI is developed for text summarization and improves the grounding of text generation by increasing the likelihood of generating tokens that are related to the text to be summarized. Differently from our method, the weight assigned to the realization of marginally likely continuations is time-independent. Similarly, Contrastive Decoding \cite{contrastive_decoding_li23, MMI_decoding_chatbots_15} has been developed to foster generations of an \textit{expert} LLM to contain plausible and fluent text with the textual input prompt by amplifying its prediction difference with respect to a weaker model. To ensure a fair comparison, we replace the text to be summarized with the input image for PMI and set the weaker model in Contrastive Decoding as the unconditioned model (see \cref{sec:other-decodings} for more details). We also compare with training-based methods that have been proposed to improve the grounding of VLMs on the visual prompt like LLaVA-RLHF \cite{RLHFouyang2022training}, Robust mPLUG-Owl \cite{liu2023aligning} and LURE \cite{zhou2023analyzing}. However, all these methods require some form of annotations, like preference data \cite{RLHFouyang2022training} or positive/negative examples \cite{liu2023aligning, zhou2023analyzing}, which M3ID+DPO does not require.

\begin{table}[t]
    \centering
        \caption{Evaluation of vision-language grounding on the validation set of MS COCO \cite{coco2014}. Captioning results are obtained by prompting the model with the task ``Describe the image''. \textit{CHAIRi} and \textit{CHAIRs }\cite{rohrbach2018object} denote the percentage of hallucinated objects and captions respectively, with lower values corresponding to fewer hallucinations. \textit{Cover} indicates the percentage of annotated objects that are mentioned in the captions. We use LLaVA-v1.3.}
\resizebox{\columnwidth}{!}{
    \begin{tabular}{lccc}
    \toprule
    & \multicolumn{3}{c}{Captioning} \\
    & CHAIRi $\downarrow$ & CHAIRs $\downarrow$ & Cover $\uparrow$ \\ \midrule
    LLaVA$_\textrm{7B}$                             &  8.1 & 17.5 & 53.3 \\
    LLaVA$_\textrm{7B}$ PMI \cite{PMI_for_summarization}                &  6.7 & 16.2 & 51.5  \\
    LLaVA$_\textrm{7B}$ Contrastive \cite{contrastive_decoding_li23}    &  6.3 & 14.8 & \textbf{55.2} \\
    \textbf{LLaVA$_\textrm{7B}$ M3ID}               &  \textbf{5.9} & \textbf{13.8} & 55.1 \\
    \midrule
    LLaVA$_\textrm{13B}$                             &  7.4 & 18.5 & \textbf{55.2} \\
    LLaVA$_\textrm{13B}$ PMI \cite{PMI_for_summarization}               &  6.5 & 16.6 & 54.7 \\
    LLaVA$_\textrm{13B}$ Contrastive \cite{contrastive_decoding_li23}  &  6.5 & 14.5 & 54.7 \\
    \textbf{LLaVA$_\textrm{13B}$ M3ID}               &  \textbf{5.5} & \textbf{13.2} & 54.0  \\
    \midrule
    LLaVA$_\textrm{13B}$ + LURE \cite{zhou2023analyzing}     & 6.4 & 27.1 & --\\
    mPLUG-Owl + LURE \cite{zhou2023analyzing}        & 5.4 & 18.8 & --\\
    \textbf{LLaVA$_\textrm{7B}$ M3ID + DPO}          & 5.7 & 13.5  & \textbf{55.8} \\
    \textbf{LLaVA$_\textrm{13B}$ M3ID + DPO}         & \textbf{5.3} & \textbf{12.6}  & 54.2 \\
    \bottomrule
    \end{tabular}
    }
    \label{tab:chair-main}
    \vspace{-0.6cm}
\end{table}

\begin{table*}[t]
    \centering
    \caption{Evaluation on the POPE VQA hallucination benchmark \cite{li2023evaluating}. $*;\dagger$ indicate results taken from \cite{liu2023aligning} and \cite{sun2023aligning} respectively. \textit{SUP.} Indicates that the method involves supervised training on annotated data, in contrast to our approaches. POPE is divided into three subsets, \textit{Random}, \textit{Popular}, and \textit{Adversarial}. We report the VQA accuracy of the model using the following template: ``Is a $\langle$object$\rangle$ present in the image?'', where we sample $\langle$object$\rangle$ randomly (Random), among the most frequent objects in the dataset (Popular), or among the objects that frequently co-occur with $\langle$object$\rangle$ (Adversarial). \textit{Acc.} is the binary classification accuracy and \textit{Yes} is the percentage of ``Yes'' answers.
    }
        \begin{tabular}{lcccccccc}
        \toprule
        &  \multicolumn{8}{c}{POPE} \\
        & \multicolumn{2}{c}{Random} & \multicolumn{2}{c}{Popular} & \multicolumn{2}{c}{Adversarial}  & \multicolumn{2}{c}{All} \\
        & Acc. $\uparrow$ & Yes (\%) & Acc. $\uparrow$ & Yes (\%) & Acc. $\uparrow$ & Yes (\%) & Acc. $\uparrow$ & Yes (\%) \\ \midrule
        Robust mPLUG-Owl-7B$^*$ \cite{liu2023aligning} $^{\rm SUP.}$        & \textbf{86.0} & --  & 73.0 & --  & 65.0 & -- & 74.7 & -- \\
        LLaVA-RLHF$_\textrm{7B}^\dagger$ \cite{sun2023aligning} $^{\rm SUP.}$ & 84.8 & 39.6 & \textbf{83.3} & 41.8 & \textbf{80.7} & 44.0 & \textbf{82.9} & 41.8  \\
        \midrule
        mPLUG-Owl-7B$^*$ \cite{ye2023mplug}                  & 52.0 & --  & 57.0 & --  & 60.0 & -- & 67.3 & -- \\
        LLaVA$_\textrm{7B}$                            & 74.8 & 75.1 & 61.8 & 86.7 & 58.1 & 90.1 & 64.9 & 84.0  \\
        \textbf{LLaVA$_\textrm{7B}$ M3ID}               & 76.0 & 67.7 & 69.3 & 73.3 & 65.8 & 77.6 & 70.3 & 72.9  \\
        \textbf{LLaVA$_\textrm{7B}$ M3ID + DPO}         & \textbf{81.2} & 65.6 & \textbf{73.9} & 67.3 & \textbf{68.2} & 75.4 & \textbf{74.4} & 69.4  \\ 
        \midrule \midrule        LLaVA-RLHF$_\textrm{13B}^\dagger$ \cite{sun2023aligning} $^{\rm SUP.}$ & \textbf{85.2} & 38.4 & \textbf{83.9} & 38.0 &\textbf{82.3} & 40.5 &\textbf{ 83.8} & 39.0  \\
        \midrule
        MiniGPT4-13B$^*$ \cite{zhu2023minigpt}               & 73.0 & --  & 67.0 & --  & 62.0 & -- & 74.7 & -- \\
        LLaVA$_\textrm{13B}$                                 & 67.9 & 80.6 & 63.8 & 83.2 & 59.8 & 87.3 & 63.8 & 83.7 \\
        \textbf{LLaVA$_\textrm{13B}$ M3ID}              & 84.3 & 55.6 & 77.0 & 61.6 & 71.3 & 68.2 & 77.5 & 61.8  \\
        \textbf{LLaVA$_\textrm{13B}$ M3ID + DPO}        & \textbf{85.2} & 53.4 & \textbf{79.1} & 57.5 & \textbf{73.2} & 67.5 & \textbf{79.2} & 51.1 \\
        \bottomrule
        \end{tabular}
        \label{tab:pope-main}
        \vspace{-0.3cm}
\end{table*}

\paragraph{Evaluation.} 
We evaluate our algorithm both on captioning and VQA benchmarks. On captioning, we measure object hallucinations by using the CHAIR (Captioning Hallucination Assessment with Image Relevance) metrics \cite{rohrbach2018object}. These metrics assess the quality of captions by comparing the mentioned objects to the annotated objects present in an image. In particular, CHAIR comprises two variants: one computes the fraction of hallucinated objects in the whole caption (CHAIRi) and the other evaluates the fraction of captions with at least one object hallucination (CHAIRs),
\begin{equation}\nonumber
    {\rm CHAIRi} = \frac{\text{\# hallucinated objects}}{\text{\# generated objects}}, 
\end{equation}
\begin{equation}\nonumber
    {\rm CHAIRs} = \frac{\text{\# hallucinated captions}}{\text{\# generated captions}}.
\end{equation}

Furthermore, we introduce the \textit{Cover} metric to quantify the comprehensiveness of the captions by measuring the fraction of ground-truth annotated objects mentioned by the model, 
\begin{equation}\nonumber
    {\rm Cover} = \frac{|\{\text{correct mentioned objects}\}|}{|\{\text{annotated objects}\}|}.
\end{equation}

To evaluate object perception, we use the POPE (Polling-based Object Probing Evaluation) VQA benchmark \cite{li2023evaluating}. In particular, POPE recasts the evaluation of object hallucinations into a binary classification problem with yes/no questions of the type ``Is a $\langle$object$\rangle$ present in the image?''. 

\paragraph{DPO training details.} 
We use Hugging Face's DPO implementation in the TRL library \cite{vonwerra2022trl} and train LLaVA for 5 epochs on 10,000 self-generated preference pairs with LoRA and cosine decaying learning rate with $2 \times 10^{-5}$ peak. Following \cite{rafailov2023direct}, we set $\beta=0.1$ in the DPO loss.

\subsection{VLM grounding on captioning}

In \cref{tab:chair-main} we test how much our method can prevent the generation of hallucinated objects in image captions. In particular, we compare against training-free and training-based baselines. We prompt all baseline methods with the instruction ``Describe the image.''~and let the models generate until the EOS token is obtained.

First, we compare M3ID with other decoding strategies, PMI \cite{PMI_for_summarization} and Contrastive Decoding \cite{contrastive_decoding_li23}. The main difference between M3ID and these baselines is that M3ID increasingly counteracts the language prior as more tokens are generated. As such, M3ID reduces ungrounded generations compared to all other training-free baselines both on the large LLaVA13B and on the smaller LLaVA7B. Furthermore, it has uniformly good improvements for different model sizes over standard multinomial decoding, in particular, M3ID achieves 27\%/21\% relative improvement over LLaVA7B and 26\%/29\% over LLaVA13B on the CHAIRi and CHAIRs metrics. Importantly, this improvement does not come at the price of high reductions of the Cover metric, which actually improves on the 7B model and decreases by less than 2.2\% for the larger 13B model\footnote{Note that CHAIR metrics can be hacked by simply returning shorter captions that do not attempt to predict any object in the image.}. Lastly, we note that M3ID shows an improvement in absolute performance that correlates with model size, suggesting that further gains could be obtained as larger models are used. In \cref{sec:longer-captioning}, we also show that M3ID still improves grounding when longer captions (twice as long on average) are generated and where other decodings do not perform as well. 

In \cref{tab:chair-main} we also report training-based results and show that pairing DPO with M3ID leads to a smaller number of hallucinated objects and improved Cover numbers compared both to our training-free approach and LLaVA+LURE \cite{zhou2023analyzing} a concurrent training based method to improve the grounding of VLMs that relies on GPT-3.5 annotations. 

\subsection{VLM grounding on VQA}

Differently from the MS COCO captioning task, the POPE VQA benchmark only requires the generation of ``Yes/No'' tokens. 
As such, one does not expect the dilution effect to play a key role. However, we highlight that the image tokens and the output of the VLM  are separated by the input template used to prime the model for the VQA task, which, as we show in \cref{sec:vqa-dilution}, introduces a non-negligible dilution effect.\footnote{We use \texttt{[Img][Model Prompt][Question]}, see \cref{sec:vqa-dilution}.} To address this, we take an offset into account when using M3ID on POPE, and select $t = t_0$ where $t_0$ is the number of tokens in between the output and the image.

In \cref{tab:pope-main}, we report the results on the POPE VQA hallucination benchmark \cite{li2023evaluating}. Hallucinations (wrong answers) tend to correlate with the tendency of the VLM to reply using the ``Yes'' token (see the disproportionately high percentage rate, 84\%/83.7\%, of ``Yes'' answers for the LLaVAv1.3 base model). This tendency can be counteracted at inference time and without any training by simply using M3ID.  In our experiments, M3ID reduces the Yes ratio to 72.9\%/61.8\% for 7B and 13B models, respectively, which leads to relative accuracy improvements over standard LLaVA decoding by 8\% and 21\%, respectively. 

To test whether our training-based approach increases the VLM's grounding on the visual prompt we trained a model with our DPO objective on the MS COCO captioning dataset and then tested it on the POPE benchmark. Note that improving caption generation does not directly imply improvements on the POPE VQA benchmark, in fact, even if the images are both from MS COCO, the output format is quite different: open-ended generation on the former and binary classification on the latter. 
However, intuitively, one expects that by improving the reliance on the visual prompt, the model could make better use of the visual information regardless of the output format required to solve the task at hand. 
\cref{tab:pope-main} shows that M3ID+DPO further improves performance over M3ID's inference time intervention. Specifically, M3ID+DPO achieves 15\%  and 24\% accuracy improvements over the LLaVA 7B and 13B models respectively. For completeness, we also compare with other training-based baselines that are fine-tuned on labeled \cite{liu2023aligning} and preference data \cite{RLHFouyang2022training}. While M3ID+DPO does not have access to any labeled information we show that it is close to these baselines without requiring any annotations.

\subsection{Ablations}\label{subsec:ablations}

\begin{figure}
    \centering
        \begin{minipage}{.5\linewidth}
          \centering
            \includegraphics[width=\linewidth]{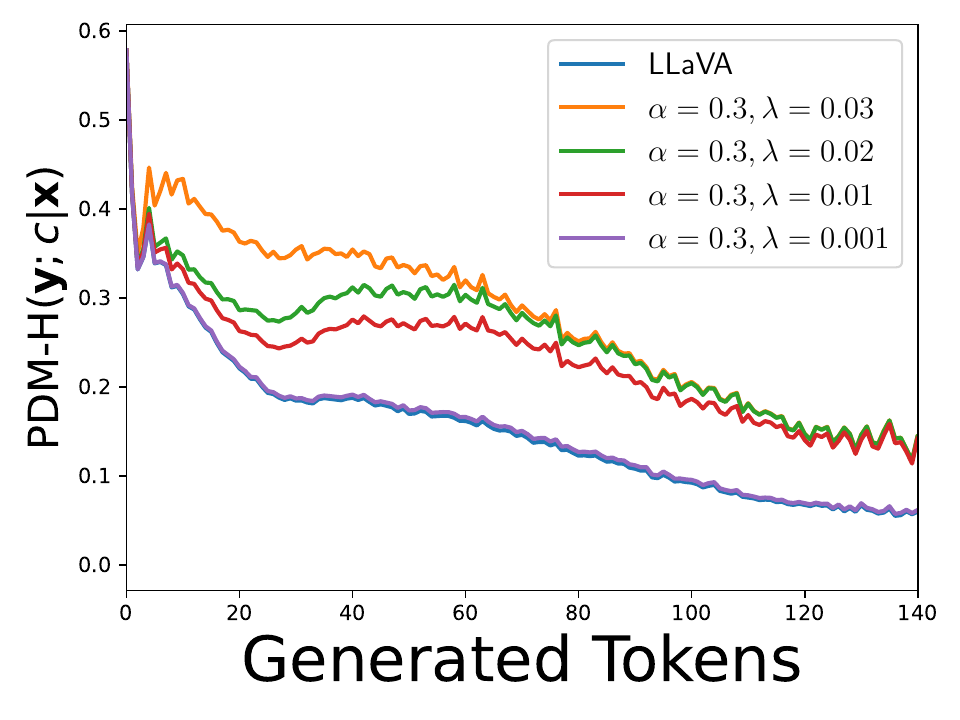}
        \end{minipage}%
        \begin{minipage}{.5\linewidth}
            \centering
            \includegraphics[width=\linewidth]{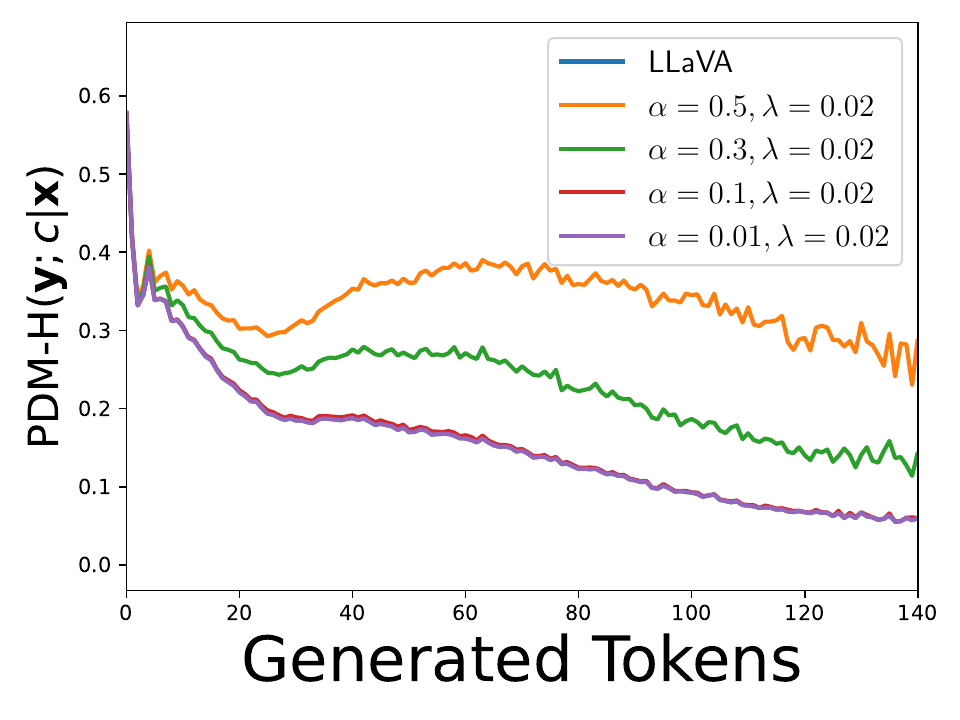}
        \end{minipage}
    \caption{\textbf{Sensitivity analysis on hyper-parameters.} We vary both the forgetting factor $\lambda$ and the thresholding parameter $\alpha$. 
    Small $\lambda$ and $\alpha$ result in minor effects. Instead, either a high forgetting factor $\lambda$ or a high $\alpha$ results in a stronger intervention which leads to higher PDM-H. 
    }
    \label{fig:sensitivity-lambda-alpha}
    \vspace{-0.5cm}
\end{figure}

\paragraph{Conditioning dilution and ``overcompensation''.}
In \cref{fig:sensitivity-lambda-alpha} and \cref{tab:sensitivity-lambda-alpha} we show that when the forgetting factor $\lambda$ is high, corresponding to the assumption that the input image gets forgotten quickly, M3ID effectively maintains a large PDM-H throughout the generation. However, while deviating from the unconditioned probability is often a desired behavior, an excessive deviation from it across the whole generation could result in ``overcompensation'', which, as we show in \cref{tab:sensitivity-lambda-alpha} can be counterproductive and can lead to higher hallucination rates. In particular, observe that higher corrections detrimentally impact the cover metric. This suggests that while M3ID tries to find tokens that diverge from the language prior the most, it is more likely to fail in captioning elements that are inherently predictable by the language prior alone. For instance, in describing \textit{an image of a dog on a leash accompanied by a man,} M3ID might overlook mentioning the presence of the man, a token that the language prior could anticipate without necessitating any visual information only from context clues. We report examples of this behavior in the \cref{sec:longer-captioning}. 

\paragraph{Importance of the confidence threshold.}
When we set a high threshold $\alpha$, the indicator function remains active more frequently along the generation, leading M3ID to consistently prioritize maximizing the distance of the generation from the language prior. Hence, similarly to the scenario with a high $\lambda$, M3ID is likely to overcompensate and increase the rate of multi-modal hallucinations. However, too high $\alpha$ disrupts linguistic fluency. We report qualitative examples of this in \cref{sec:longer-captioning}. In \cref{tab:ablation-studies} we also show that adding the term to compensate for contextual pressure does not significantly change the results over LLaVA decoding in the POPE benchmark since it is essentially framed as a binary classification over the ``Yes''/ ``No'' tokens and not open-ended generation like captioning.

\begin{table}[t]
    \centering
        \caption{\textbf{Sensitivity to hyper-parameters.} $\lambda$ is the forgetting factor used to counteract conditioning dilution while $\alpha$ is the threshold used to mitigate overcompensation when the contextual pressure is high. We report results on LLaVA-7B.    \vspace{-0.1cm}}
\resizebox{\columnwidth}{!}{
    \begin{tabular}{lccc}
    \toprule
    & \multicolumn{3}{c}{Captioning} \\
    & CHAIRi $\downarrow$ & CHAIRs $\downarrow$ & Cover $\uparrow$ \\ \midrule
    LLaVA$_\textrm{7B}$                                 &  8.0 & 17.8 & 53.5 \\ 
    \textbf{M3ID} $\alpha=0.3, \lambda = 0.02$          &  \textbf{5.8} & \textbf{13.6} & \textbf{55.0} \\
    \midrule
    M3ID $\alpha=0.3, \lambda = 0.001$         &  7.3 & 17.4 & 53.7 \\
    M3ID $\alpha=0.3, \lambda = 0.03$          &  6.2 & 14.7 & 52.0 \\
    M3ID $\alpha=0.3, \lambda = 0.1$           &  7.2 & 16.3 & 45.5 \\
    \midrule
    M3ID $\alpha=0.5, \lambda = 0.02$          &  6.1 & 14.9 & 54.6 \\
    M3ID $\alpha=0.1, \lambda = 0.02$          &  6.9 & 15.4 & 53.1 \\
    M3ID $\alpha=0.01, \lambda = 0.02$         &  6.9 & 16.6 & 52.5 \\
    \bottomrule
    \end{tabular}
    }
    \label{tab:sensitivity-lambda-alpha}
    \vspace{-0.1cm}
\end{table}

\begin{table}
    \centering
    \caption{\textbf{Ablation studies.} We ablate the components of M3ID on both model sizes. Removing the amplification term to counteract conditioning dilution leads to more hallucinations. \vspace{-0.1cm}}
\resizebox{\columnwidth}{!}{
    \begin{tabular}{lcccc}
    \toprule
    & POPE & \multicolumn{3}{c}{{Captioning}} \\
    & {Accuracy $\uparrow$} & {CHAIRi} $\downarrow$ & {CHAIRs} $\downarrow$ & {Cover} $\uparrow$ \\ \midrule
    LLaVA$_\textrm{7B}$                   & 64.9 & 8.0 & 17.8  & 53.5  \\
    w/ context pressure                  & 65.5 & 6.9 & 16.7 & {54.5}   \\ 
    w/ conditioning dilution             & \textbf{77.5} & 6.4 & 14.1 & {53.9}   \\  
    \textbf{LLaVA$_\textrm{7B}$ M3ID}     & \textbf{77.5} & \textbf{5.8} & \textbf{13.6} & \textbf{55.0}   \\ 
    \midrule
    LLaVA$_\textrm{13B}$                  & 63.8 & 15.3 & 18.2 & \textbf{55.3}  \\
    w/ context pressure                  & 64.9 & 6.5 & 14.4 & 55.0  \\ 
    w/ conditioning dilution             & \textbf{70.3} & 5.9 & 14.8 & 53.7  \\ 
    \textbf{LLaVA$_\textrm{13B}$ M3ID}    & \textbf{70.3} & \textbf{5.4} & \textbf{13.0} & 54.0  \\
    \bottomrule
    \label{tab:ablation-studies}
    \end{tabular}
}
\vspace{-0.8cm}
\end{table}

\section{Conclusions}

We introduced M3ID, a new approach designed to combat multi-modal hallucinations by maximizing the mutual information between the text generated by VLMs and the corresponding visual context. M3ID operates at inference time and can be seamlessly integrated with any pre-trained autoregressive VLM. This makes M3ID a cost-effective and flexible solution to enhance vision-language grounding.
Furthermore, for settings where model training is feasible and higher visual grounding is expected, we also paired M3ID with Direct Preference Optimization (DPO), showcasing reduced hallucinatory behaviors. 
Interestingly, our findings suggest that object hallucinations in VLMs result from excessive reliance on the language prior rather than a poor understanding of the visual modality. In fact, M3ID at inference time is sufficient to significantly reduce the amount of generated hallucinations without any training.

A limitation of M3ID is that it requires two forward passes at inference time, one for the conditioned and one for the unconditioned prediction. 
A possible solution to not increase inference time, but at the expense of higher memory consumption, is to use two batched queries with one having masked visual tokens. Furthermore, as we observed in \cref{tab:chair-main}, sometimes M3ID may prevent the generation of objects that are highly likely under the unprompted language prior (due to context clues).
Interestingly, a similar observation has been reported in \cite{Spain2011-perona}, where, people asked to provide a 10-word list of objects contained in a given image often failed to report the most obvious objects while mainly focusing on secondary ones.
While we already showed that this can be mitigated with proper hyper-parameter selection \cref{tab:sensitivity-lambda-alpha} we believe that an interesting avenue for future research is to directly encode this preference within the model's weights by favoring the generation of structured captions that get progressively more detailed while still being grounded to the image.

In general, integrating human- or AI-annotated preference pairs, assessed based on their level of grounding, constitutes a promising avenue for future investigation.

{
    \small
    \bibliographystyle{ieeenat_fullname}
    \bibliography{main}
}

\clearpage
\renewcommand{\theHsection}{A\arabic{section}}
\renewcommand{\thesection}{\Alph{section}}
\renewcommand{\thesubsection}{\Alph{section}.\arabic{subsection}}
\setcounter{section}{0}

\maketitlesupplementary

\section{Experimental details}\label{sec:experimental_details}

In this section, we provide details on our experimental setup.

\paragraph{Computing CHAIR metrics.} We select 5,000 images from the 2014 validation split of MS COCO \cite{coco2014} and, for each image, prompt the models with one of the following questions: ``Describe the image.'', ``Give an explanation of the image.'', ``Provide a description of the given image.''. To measure hallucinated objects we follow \cite{rohrbach2018object} and complement the set of MS COCO annotated objects with their synonyms and automatically detect object hallucinations by comparing the objects mentioned in the captions against the annotated ones.

\paragraph{Computing POPE metrics.} We use the official sets of questions introduced in \cite{li2023evaluating}, consisting of 3,000 \textit{random}, 3,000 \textit{popular}, and 3,000 \textit{adversarial} questions on images taken from the 2014 validation split of MS COCO \cite{coco2014}. We report the VQA accuracy using the following template: ``Is a $\langle$object$\rangle$ present in the image?'', where we sample $\langle$object$\rangle$ randomly (Random), among the most frequent objects in the dataset (Popular), or among the objects that frequently co-occur with $\langle$object$\rangle$ (Adversarial). 

\paragraph{Decoding hyper-parameters.} To find the optimal hyper-parameters for M3ID, PMI, and Contrastive Decoding we compute the CHAIR metrics on 500 images sampled from the MS COCO validation set that do not overlap with the 5,000 images used to measure the CHAIR metrics. For M3ID, we search $\alpha$ in the range $\{0, 0.01, 0.1, 0.3, 0.5, 0.8, 1.0\}$ and the optimal fading memory coefficient $\lambda$ in $\{0.001, 0.005, 0.01, 0.02, 0.03\}$.
For PMI \cite{PMI_for_summarization}, we search $\tau$ in $\{0, 0.01, 0.1, 0.3, 0.5, 0.8, 1.0\}$ and $\mu$ in $\{0.1, 0.3, 0.5, 0.8, 1.5, 2.0\}$. For Contrastive Decoding \cite{contrastive_decoding_li23}, we search $\xi$ in $\{0.1, 0.3, 0.5, 0.8, 1.5, 2.0\}$ and $\psi$ in $\{0, 0.01, 0.1, 0.3, 0.5, 0.8, 1.0\}$.

\paragraph{DPO hyper-parameters.} We use the same set of hyper-parameters as the ones used to fine-tune LLaVA with SFT \cite{llava}. In particular, we train using the AdamW optimizer with batch size 128, learning rate $2 \times 10^{-5}$, cosine annealing scheduler, warm-up ratio 0.03, and DeepSpeed ZeRO-2. We use LoRA with rank 64, scaling factor 16, and dropout 0.05. Following \cite{rafailov2023direct}, we set $\beta=0.1$ in the DPO loss.

\paragraph{Hardware.} Experiments are run on 8 NVIDIA Tesla A100 GPUs.

\section{Context-dependent decodings}\label{sec:other-decodings}

In this section, we review precedent context-dependent decoding strategies developed for large language models that we use as baselines in our experiments and highlight the differences with M3ID.

\paragraph{PMI Decoding.} Pointwise Mutual Information (PMI) Decoding \cite{PMI_for_summarization} is a decoding strategy developed for improving the grounding of summary generation by increasing the likelihood of generating tokens that are related to the text to be summarized. Specifically, PMI optimizes for the mutual information of the source text and target token when the model exhibits uncertainty -- quantified with the Shannon entropy of the output distribution. Let $p$ denote an LLM, $\x$ a prompt and $c$ the source text to be summarized, PMI selects tokens as follows,
\begin{equation*}
    y_t = \arg \max_{y \in \mathcal{V}} \log p(y|\y_{<t},\x,c) - \mu \mathbbm{1} \left[ H(p(\cdot|\y_{<t},\x,c)) \geq \tau \right] \log p(y|\y_{<t},\x)
\end{equation*}
where $H(p(\cdot|\y_{<t},\x,c)) = - \sum_{y \in \mathcal{V}} p(y|\y_{<t},\x,c) \log p(y|\y_{<t},\x,c)$ denotes the Shannon entropy. 
In the multi-modal case, we replace the text to be summarized with the input image. Notice that, differently from our method, in the PMI intervention the weight $\mu$ assigned to the realization of marginally likely continuations is constant across time. 

\paragraph{Contrastive Decoding.} Contrastive Decoding \cite{contrastive_decoding_li23, MMI_decoding_chatbots_15} has been developed to foster generations of an \textit{expert} LLM $p_{\rm exp}$ to contain plausible and fluent text with the textual input prompt by amplifying its prediction difference with respect to a weaker \textit{amateur} model $p_{\rm ama}$. Firstly, Contrastive Decoding applies a \textit{plausibility constraint} in order to mask tokens to which the expert model assigns low probability, \ie,
\begin{equation*}
    \mathcal{V}_{\rm plausible} = \left\{ v \in \mathcal{V}, \; (\log p_{\rm exp})_v \geq \log \psi + \max_{k \in \mathcal{V}} (\log p_{\rm exp})_k \right\}. 
\end{equation*}
Secondly, Contrastive Decoding applies a penalty to the amateur logits,
\begin{equation*}
    y_t = \arg \max_{y \in \mathcal{V}_{\rm plausible}} (1+\xi)\log p_{\rm exp}(y|\y_{<t}) - \xi \log p_{\rm ama}(y|\y_{<t}).
\end{equation*}
In the multi-modal case, we set the expert model as the VLM, \ie, $p_{\rm exp}(\cdot) = p(\cdot|\x,c)$ and the amateur model as the unconditioned model, \ie, $p_{\rm ama}(\cdot) = p(\cdot|\x)$. Notice that, also in this case, the penalization is time-independent.

\paragraph{Multi-Modal Mutual-Information Decoding (M3ID).}
Our decoding algorithm M3ID optimizes the vision-language grounding of VLM generations by maximizing the mutual information between the generated textual tokens $\y$ and the provided visual context $c$. Specifically, when the contextual pressure is low (see \cref{sec:methods}), M3ID applies a penalty to the log probabilities unconditioned on the image, \ie, it penalizes the language-only prior of the VLM. In contrast with the previous strategies and motivated by our fading-memory modeling assumption, in M3ID the strength of this penalization increases as more tokens are generated and it is controlled by the time-dependent parameter $\gamma_t$. Let $p$ denote the VLM and $\x$ a textual prompt, M3ID selects new tokens as follows,
\begin{equation*}
    \begin{cases}
            {\displaystyle y_t = \arg \max_{y \in \mathcal{V}} \log p(y|\y_{<t},\x,c) - \mathbbm{1} \left[\max_k ( p(\cdot|\y_{<t},\x,c))_k < \alpha \right] \frac{1-\gamma_t}{\gamma_t} \left(\log p(y|\y_{<t},\x,c) - \log p(y|\y_{<t},\x) \right)} \\
            {\displaystyle \gamma_t = e^{-\lambda(t+t_0)}},
    \end{cases}
\end{equation*}
where the parameter $\lambda$ controls the decay rate of the fading memory and $t_0$ controls the strength of the language-prior penalization at the beginning of the generation. In all our captioning experiments, we set $t_0=0$, whereas when evaluating on POPE we find that using $t_0=10$, approximately equal to the length of the question $\x$ which separates the image $c$ and the beginning of the answer $\y$, improves the results (see also \cref{sec:vqa-dilution}). 

\section{Multi-modal Direct Preference Optimization} 

In this section, we present the theoretical formulation of multi-modal Direct Preference Optimization (DPO) and our algorithm for aligning VLMs on self-generated data.

\paragraph{Theoretical formulation.} Given an image context $c$, let $\y_w$ and $\y_l$ be two different continuations to the same prompt $\x$, with $\y_w$ preferred over $\y_l$ ($\y_w \succcurlyeq \y_l$) judging based on grounding with respect to $c$. Consider the common assumption that the preference is governed by a latent Bradley-Terry preference model \cite{bradley1952rank} with the reward given by $r^*(c,\x,\y)$ and higher reward corresponding to better vision-language grounding. Thus, the preference distribution can be written as a sigmoid of the rewards' difference,
\begin{equation}\nonumber
    p^*(\y_w \succcurlyeq \y_l |c,\x) = \sigma\left(r^*(c,\x,\y_w) - r^*(c,\x,\y_l)\right).
\end{equation}
The preference optimization objective is to find the optimal policy $\hat{p}^*$ that maximizes the expected reward of the generated continuations and minimizes the KL divergence with the reference policy $p_{\rm ref}$, which is the policy at initialization\footnote{This regularization term is often added to avoid reward hacking.},
\begin{equation}\nonumber
    \hat{p}^* = \arg\max_p \mathbb{E}_p \, r^*(c,\x,\y) - \beta D_{\rm KL}(p,p_{\rm ref}),
\end{equation}
where $\y \sim p(\,\cdot\,|c,\x)$. Leveraging an analytical mapping between $r^*$ and $p^*$, Direct Preference Optimization (DPO) \cite{rafailov2023direct} allows to directly optimize the policy on a preference dataset $\mathcal{D} = \{(c^i,\x^i,\y_w^i,\y_l^i)\}_i$ using the loss
\begin{align}\label{eq:dpo}
    \mathcal{L}_{\rm DPO} = - \mathbb{E}_{(c,\x,\y_w,\y_l)\sim \mathcal{D}} \left[ \log \sigma \left( \beta \log \frac{p(\y_w|c,\x)}{p_{\rm ref}(\y_w|c,\x)} - \beta \log \frac{p(\y_l|c,\x)}{p_{\rm ref}(\y_l|c,\x)} \right)\right]. \nonumber
\end{align}
Optimizing for this loss increases the likelihood of the preferred and better-grounded completions $\y_w$ and decreases the likelihood of the poorly-grounded completions $\y_l$.

\paragraph{Preference data generation and alignment.} Our alignment procedure is as follows.
\begin{enumerate}
    \item We start from $n$ images $\{c^i\}_{i\in [n]}$ and obtain one positive caption $\y_w^i$ per image prompting a VLM with the task $\x^i=\,$``Describe this image.'' In order to maximize vision-language grounding, we use M3ID to generate the answers.
    \item To generate the negative captions $\{\y_l^i\}_{i\in [n]}$, we use the same set of images and prompt an unconditioned VLM (\ie, the VLM with masked visual tokens) to complete the first sentences generated for the positive captions. In such a way, the first sentence is well-grounded in the image content, while the continuations are dictated by the language prior of the VLM alone and serve as informative negative examples.
    \item We train the VLM on the self-generated preference dataset $\mathcal{D} = \{(c^i,\x^i,\y_w^i,\y_l^i)\}_i$ with the DPO loss and the hyper-parameters described in \cref{sec:experimental_details}.
\end{enumerate}

\section{Further results}

In this section, we present further results complementing the experiments presented in the main text.

\begin{figure}
    \centering
    \begin{minipage}{.35\linewidth}
        \centering
        \includegraphics[width=\linewidth]{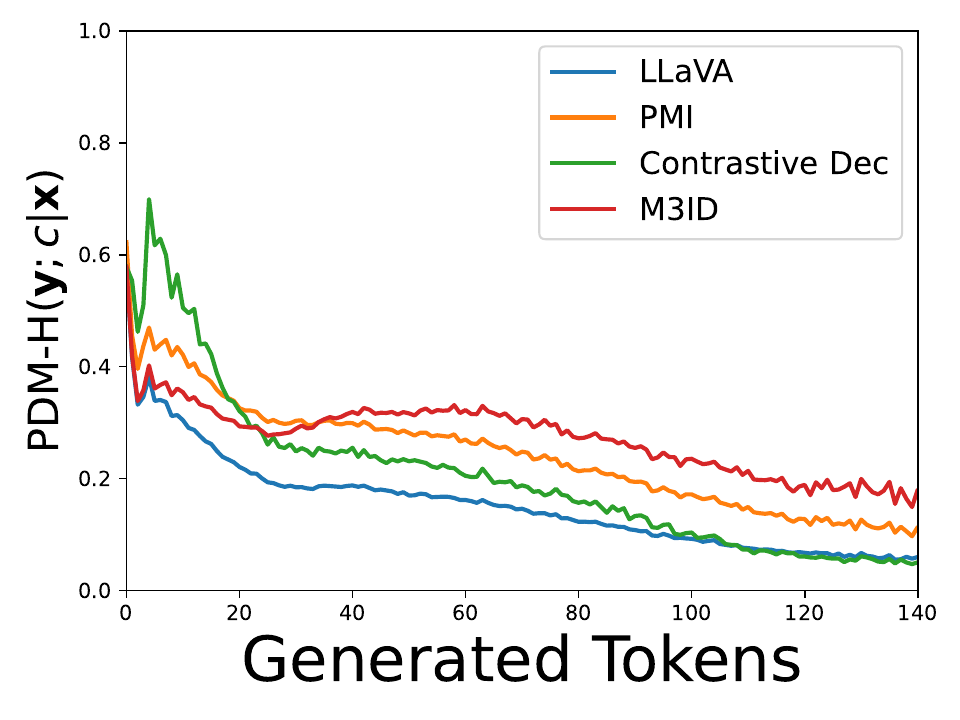}
    \end{minipage}%
    \hspace{1cm}
    \begin{minipage}{.35\linewidth}
        \centering
        \includegraphics[width=\linewidth]{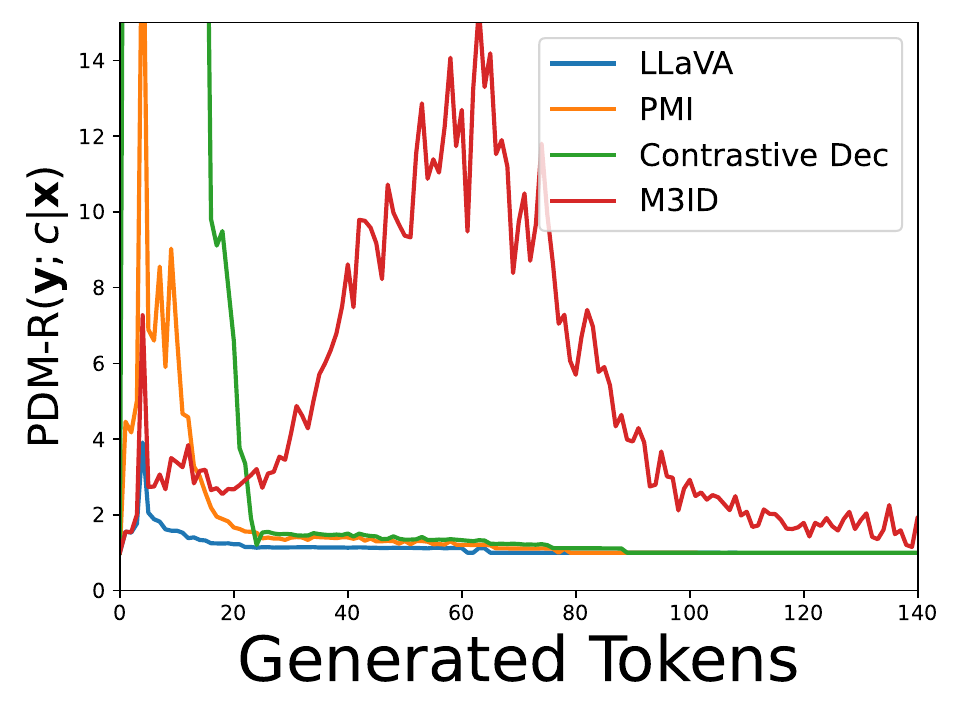}
    \end{minipage}
    \caption{\textbf{PDMs with different distances.} We report how PDM varies with different choices of the distance function, \ie, PDM-H with Hellinger (left panel) and PDM-R with Rank (right panel). We compare the generation of a caption with M3ID, standard, and other context-aware decoding schemes. With both distance functions, M3ID maximizes the separation of the conditioned and unconditioned distribution, effectively counteracting the language prior even throughout long caption generations.}
    \label{fig:different-distances}
\end{figure}

\begin{figure}[b]
    \centering
    \begin{minipage}{.49\linewidth}
      \centering
    \includegraphics[width=\linewidth]
    {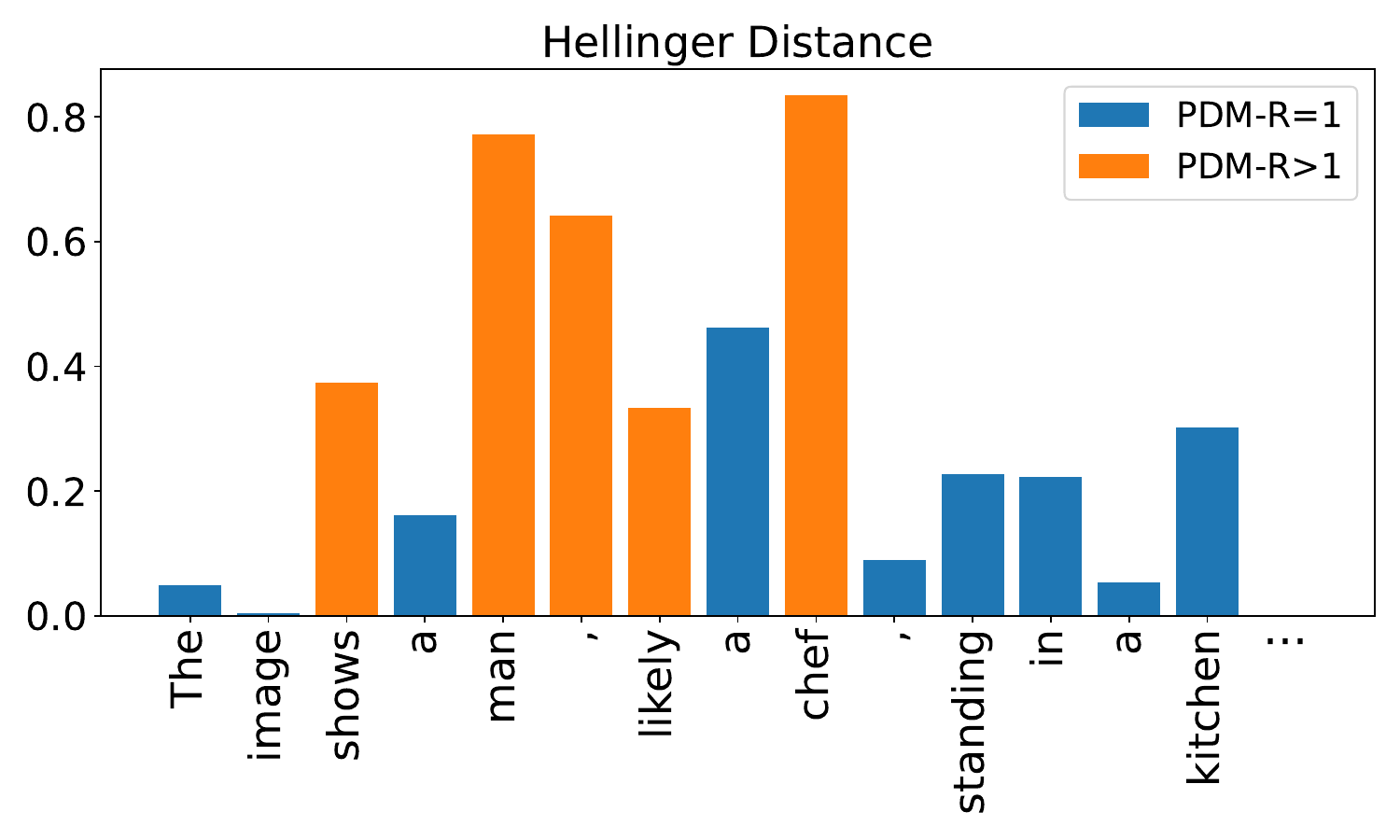}
    \end{minipage}
    \hspace{.5cm}
    \begin{minipage}{.45\linewidth}
        \centering
    \includegraphics[width=\linewidth]{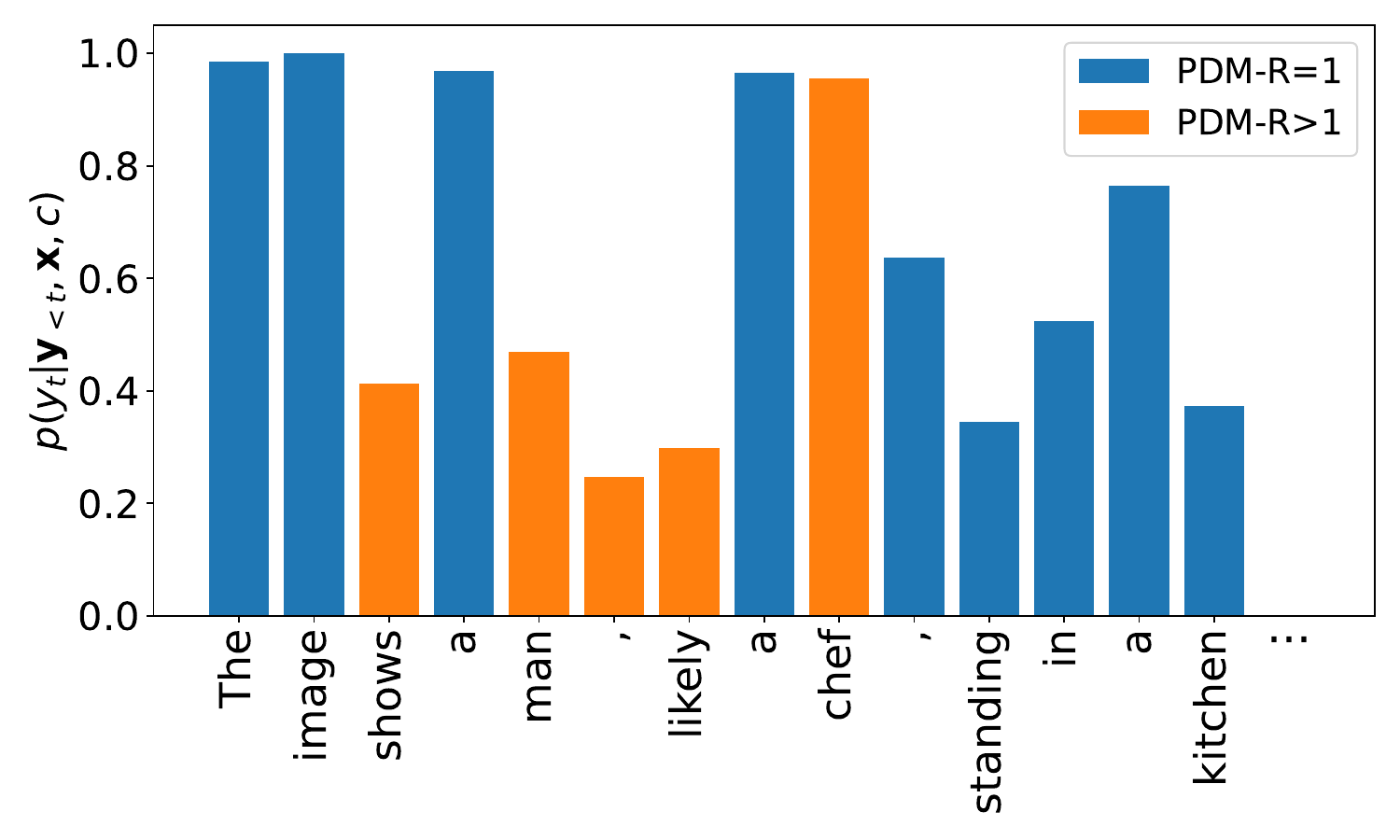}
    \end{minipage}
    \caption{\textbf{How much does the conditioning prompt ``surprise'' the unconditioned predictor?} We report the PDM-H and the PDM-R for each token in the string ``The image shows ...''. Both distances increase on tokens for which visual information is required and decrease for punctuation, articles, and when sufficient visual information has already been extracted and is present in the caption (\eg, ``kitchen'' is already very likely given the text alone). 
    }
    \label{fig:surprise-appendix}
\end{figure}

\subsection{Prompt dependency measures}\label{sec:PDM-appendix}

\paragraph{Impact of different distance functions.} Depending on the choice of the distance function, PDMs highlight different aspects of the generative distribution. 
For example, by choosing the Hellinger ($\text{PDM-H}$) distance, we consider the whole generative distribution over the vocabulary tokens. However, notice that the generative process is mainly determined by its high probability modes, especially when greedy decoding or low-temperature sampling is used. Hence $\text{PDM-H}$ relies on a high number of irrelevant tokens. To account for this, we complement $\text{PDM-H}$ with the following PDM which only depends on the model's preference in generating the most likely token: 
\begin{equation}
    \text{PDM-R}(\y, c| \x) \triangleq {\rm rank}_{p(\y|\x)}(\arg\max {\rm rank}_{p(\y|\x, c)})
\end{equation}
where ${\rm rank}_p$ is the ranking of the tokens in the vocabulary according to the distribution $p$. Note that when $\text{PDM-R}=1$, the highest-ranking token with the conditioning image is also the highest-ranking token without the image. So a greedy generation without the specific conditioning signal would result in the same continuation.

In \cref{fig:different-distances}, we show how the average $\text{PDM-H}$ and $\text{PDM-R}$ vary during the generation of 5000 captions using LLaVA's standard decoding, M3ID, and the context-dependent decoding baselines.
PDMs with standard decoding decrease as more tokens are generated in both cases. In particular, we observe that after approximately 30 tokens have been generated, $\text{PDM-R}$ approaches 1, signaling that, on average, the VLM effectively behaves as an LLM which is unconditioned on the visual prompt. PMI and Contrastive Decoding are effective in increasing the PDM distance at the beginning of the captions but their effect becomes negligible near the middle and end of the generation, when more multi-modal hallucinations are observed. In contrast, M3ID maximizes both PDMs and successfully counteracts the language prior even when generating longer captions. Notice that $\text{PDM-R}$ has a spike near the center of the generated captions signaling that M3ID selects, on average, tokens that are ranked among the top 15 by the model without conditioning. We attribute this phenomenon to the fact that there exist several ways to continue the captions after the beginning and before the end of the captions, which are more ``constrained'' (\eg, the majority of captions start as ``In the image there are ...'').

\paragraph{Decay rate of conditioning dilution.} 
In \cref{fig:fading-memory} and \cref{fig:different-distances} we report PDMs for the LLaVA model. Note that to reduce the computational complexity of the hyper-parameter search of our method it is possible to estimate the decay rate $\lambda$ directly from these plots. 
In fact, estimating the decay rate from \cref{fig:fading-memory} with linear regression in logarithmic coordinates, leads to $\lambda=0.016$, which is close to the optimal value 0.02 found with cross-validation.

\paragraph{Surprising the unconditioned model.} \cref{fig:surprise-appendix} shows the PDM-H and the PDM-R for each token in the string ``The image shows a man, likely a chef, standing in a kitchen ...''. We color code the bars according to PDM-R, in blue when the conditioned and unconditioned models share the same prediction and in orange when they differ.
Note that both distances increase on tokens for which visual information is required (\eg objects and attributes), while they decrease for articles. Also, note that both models mostly agree even when sufficient visual information has already been extracted and is present in the caption (\eg, ``kitchen'' is already very likely given the text alone). On the right of \cref{fig:surprise-appendix}, we show that tokens like ``the'' and ``a'' have high probability according to the VLM and have $\text{PDM-R}=1$.

\subsection{Captioning}\label{sec:longer-captioning}

\paragraph{Detailed captioning and stochasticity due to sampling.} \cref{tab:chair-detailed} presents the CHAIR and cover metrics for detailed captioning, where LLaVA is prompted to generate longer captions resulting in an increase of the cover metric from approximately 55\% to about 70\%. In this scenario, although all methods exhibit larger CHAIR values compared to standard captioning tasks, M3ID and M3ID+DPO achieve the best results in minimizing object hallucinations, substantiating the effectiveness of our approaches. Additionally, this table includes the standard deviations resulting from stochastic sampling, which we omitted in the main text for the sake of clarity.

\begin{table*}[t]
    \centering
        \caption{Evaluation of vision-language grounding on the validation set of MS COCO \cite{coco2014}. Captioning results have been obtained by prompting the model with the task ``Describe the image'', and detailed captioning results with the task ``Describe the image in detail.'' \textit{CHAIRi} and \textit{CHAIRs }\cite{rohrbach2018object} denote the percentage of hallucinated objects and captions respectively, with lower values corresponding to fewer hallucinations. \textit{Cover} indicates the percentage of annotated objects that are mentioned in the captions. 
        }
    \begin{tabular}{lcccccc}
    \toprule
    & \multicolumn{3}{c}{Captioning} & \multicolumn{3}{c}{Detailed Captioning} \\
    & CHAIRi $\downarrow$ & CHAIRs $\downarrow$ & Cover $\uparrow$ & CHAIRi $\downarrow$ & CHAIRs $\downarrow$ & Cover $\uparrow$ \\ \midrule
    LLaVA$_\textrm{7B}$ &  8.1 {\small $\pm$ 0.5} & 17.5 {\small $\pm$ 0.8} & 53.3 {\small $\pm$ 0.7} & 10.1 {\small $\pm$ 0.3} & 35.6 {\small $\pm$ 0.7} & \textbf{71.1} {\small $\pm$ 0.5}  \\
    LLaVA$_\textrm{7B}$ PMI \cite{PMI_for_summarization}   &  6.7 {\small $\pm$ 0.7} & 16.2 {\small $\pm$ 1.2} & 51.5 {\small  $\pm$ 0.6} & 9.8 {\small $\pm$ 0.4} & 34.0 {\small $\pm$ 0.9} & 70.1 {\small $\pm$ 1.1} \\
    LLaVA$_\textrm{7B}$ Contrastive \cite{contrastive_decoding_li23}               &  6.3 {\small $\pm$ 0.1} & 14.8 {\small $\pm$ 0.1} & 55.2 {\small $\pm$ 0.1} & 10.1 {\small $\pm$ 0.1} & 33.7 {\small $\pm$ 0.3} & 69.9 {\small $\pm$ 0.1} \\
    \textbf{LLaVA$_\textrm{7B}$ M3ID} &  5.9 {\small $\pm$ 0.4} & 13.8 {\small $\pm$ 0.8}  & 55.1 {\small  $\pm$ 0.4} & 8.8 {\small $\pm$ 0.4} & 28.8 {\small $\pm$ 0.8} & 67.4  {\small $\pm$ 0.7} \\
    \textbf{LLaVA$_\textrm{7B}$ M3ID + DPO} & \textbf{5.7} {\small $\pm$ 0.4} & \textbf{13.5} {\small $\pm$ 0.7} & \textbf{55.8} {\small $\pm$ 0.2} & \textbf{8.5} {\small $\pm$ 0.6} & \textbf{28.1} {\small $\pm$ 0.6} & 68.9 {\small $\pm$ 0.9} \\
    \midrule
    LLaVA$_\textrm{13B}$ &  7.4 {\small $\pm$ 0.4} & 18.5 {\small $\pm$ 1.0} & \textbf{55.2} {\small $\pm$ 0.4} & 8.5 {\small $\pm$ 0.5} & 31.8 {\small $\pm$ 1.7} & 67.0 {\small $\pm$ 0.7} \\
    LLaVA$_\textrm{13B}$ PMI \cite{PMI_for_summarization}  &  6.5 {\small $\pm$ 0.5} & 16.6 {\small $\pm$ 1.3} & 54.7 {\small $\pm$ 0.6} & 8.6 {\small $\pm$ 0.3} & 28.5 {\small $\pm$ 1.5} & 67.1 {\small $\pm$ 0.7} \\
    LLaVA$_\textrm{13B}$ Contrastive \cite{contrastive_decoding_li23}  &  6.5 {\small $\pm$ 0.1} & 14.5 {\small $\pm$ 0.2} & 54.7 {\small $\pm$ 0.1} & 8.1 {\small $\pm$ 0.2} & 28.7 {\small $\pm$ 0.4} & \textbf{68.2} {\small $\pm$ 0.2} \\
    \textbf{LLaVA$_\textrm{13B}$ M3ID} & 5.5 {\small $\pm$ 0.3} & 13.2 {\small $\pm$ 0.8} & 54.0 {\small $\pm$ 0.4} &  7.8 {\small $\pm$ 0.6} & 27.5 {\small $\pm$ 0.8} & \textbf{68.2} {\small $\pm$ 0.6} \\
    \textbf{LLaVA$_\textrm{13B}$ M3ID + DPO} & \textbf{5.3} {\small $\pm$ 0.2} & \textbf{12.6} {\small $\pm$ 0.7}  & 54.2 {\small $\pm$ 0.3} & \textbf{7.5} {\small $\pm$ 0.4}& \textbf{26.9} {\small $\pm$ 0.7} & 67.7 {\small $\pm$ 0.5} \\
    \bottomrule
    \end{tabular}
    \label{tab:chair-detailed}
\end{table*}

\begin{table}[t]
    \centering
    \caption{\textbf{M3ID introduced error metrics.} Frequency at which M3ID either modifies or maintains the hallucinations that LLaVA produces in the task of captioning MS COCO images \cite{coco2014} (same setting as in \cref{tab:chair-main}).}
    \begin{tabular}{lcccc}
    \toprule
    LLaVA &  correct &  correct  &  incorrect  &  incorrect  \\
    M3ID  &  correct &  incorrect  &  correct  &  incorrect  \\ \midrule
    LLaVA$_\textrm{7B}$                             & 72.0 \% & 3.0 \% & 16.2 \%  & 9.8 \% \\
    LLaVA$_\textrm{13B}$                            & 74.2 \% & 2.4 \% & 18 \% & 5.4 \% \\
    \bottomrule
    \end{tabular}
    \label{tab:introduced-errors}
\end{table}

\paragraph{Introduced errors.} To further illustrate the effectiveness of M3ID in reducing hallucinations, in \cref{tab:introduced-errors} we reconsider the metrics presented in \cref{tab:chair-main} before aggregration. Specifically, we detail the proportion of hallucination-free captions generated by both the base model and M3ID, alongside the proportion of captions that contain hallucinations exclusively in one of the two cases, and the proportion of captions where hallucinations occur under both models. For the 13B model (respectively 7B model), M3ID corrects hallucinations in 18\% (16.2\%) of instances, while introducing new hallucinations in only 2.4\% (3.0\%) of cases.

\begin{table}[t]
    \caption{\textbf{Examples of overcompensation.} Maximally surprising the VLM might lead to ``overcompensation'', a phenomenon affecting decoding interventions that amplify the logits difference between the conditioned and unconditioned models like M3ID.
    When M3ID's hyper-parameters force a strong correction over the language prior, the resulting captions tend to overlook elements that are inherently predictable by the language prior based on the text tokens alone, \eg, the man in the left picture and the van in the right picture.}
    \centering
    \begin{tabular}{p{0.16\linewidth} p{0.38\linewidth} p{0.38\linewidth}}
    \toprule
    Input Image & 
    \includegraphics[width=.9\linewidth]{figures/man-and-dog-overcompensation.png} 
    & 
    \includegraphics[width=.78\linewidth]{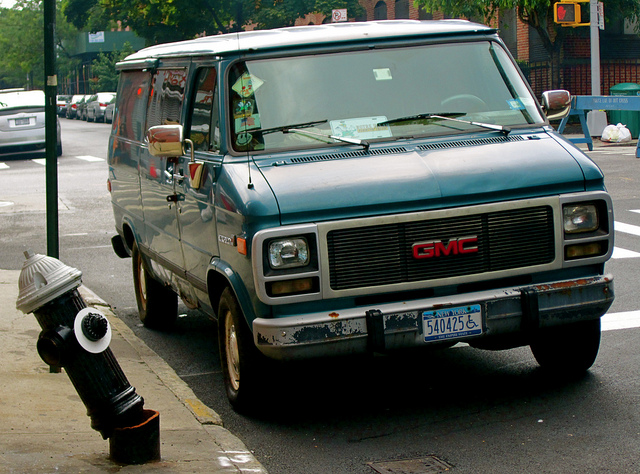} 
    \\ 
    \midrule
    Input Instruction & \multicolumn{2}{c}{\textit{Describe the image.}} \\
    \midrule
    \midrule
    LLaVA$_\textrm{7B}$           
    & The image shows a man walking a dog on a leash, with the dog wearing a raincoat. They are walking down a sidewalk, and the man is holding an umbrella to protect himself from the rain. 
    &  The image shows a blue van parked on the side of a street, with a fire hydrant nearby. The van is parked on the side of the road, and there is a fire hydrant located close to it.  \\
    \midrule
    LLaVA$_\textrm{7B}$ overcompensation     
    & The image features a dog on a leash, walking down a paved road or pathway. [\textit{No man is mentioned}]
    & The image features a city street with a sidewalk, a fire hydrant, and a pole. There is also a street sign visible in the scene. [\textit{No van is mentioned}]
    \\
    \midrule
    LLaVA$_\textrm{7B}$ M3ID      
    & The image shows a man walking a dog on a leash in a city street. The man is holding an umbrella over the dog to protect it from rain. They are walking along a dirt road near mountains. 
    & The image features a blue van parked on the side of a city street next to a fire hydrant and on the curb.  
    \\
    \bottomrule
    \end{tabular}
    \label{tab:overcompensation}
\end{table}

\paragraph{Overcompensation.} \cref{tab:overcompensation} illustrates instances of ``overcompensation'', a phenomenon affecting decoding interventions that amplify the logits difference between the conditioned and unconditioned models.
For example, M3ID, while trying to select tokens that deviate from the language prior, tends to overlook elements that are inherently predictable by the language prior alone (see \cref{subsec:ablations}). In describing the image of a dog on a leash with a man (left panel in \cref{tab:overcompensation}), when M3ID is applied with an excessively large penalty applied to the language prior, it fails to mention the presence of the man. This oversight occurs because the ``man'' token could be inferred from context clues (first tokens in the caption) by the language prior alone without necessitating any visual information. A similar behavior is observed in the second example (right panel), where M3ID fails to mention the presence of the van, despite it being the dominant element in the image.

\begin{table}[t]
    \caption{\textbf{Examples of poor language fluency.} Using a high threshold value $\alpha$ ($\alpha=1$ in the examples) disrupts language fluency. Intensifying the strength of the correction term results in the model initially losing syntactical accuracy and subsequently generating apparently random tokens, including non-English ones.}
    \centering
    \begin{tabular}{p{0.16\linewidth} p{0.38\linewidth} p{0.38\linewidth}}
    \toprule
    Input Image & 
    \includegraphics[width=.83\linewidth]{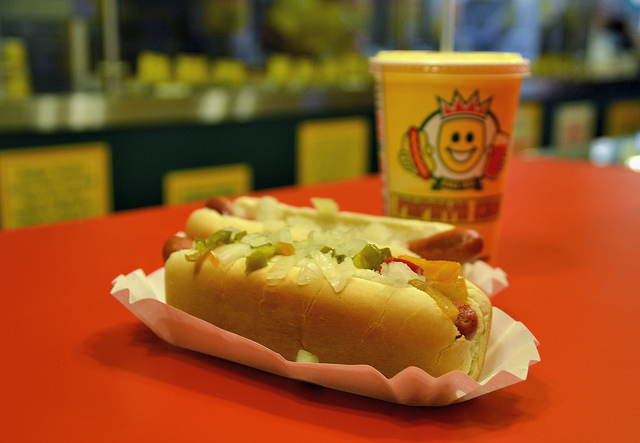} 
    & \includegraphics[width=.78\linewidth]{figures/van-overcompensation-ablation.png} 
    \\ 
    \midrule
    Input Instruction & \multicolumn{2}{c}{\textit{Describe the image.}} \\
    \midrule
    \midrule
    LLaVA$_\textrm{7B}$           
    & The image shows a hot dog on a bun, accompanied by a cup of soda, sitting on a red table.  
    &  The image shows a blue van parked on the side of a street, with a fire hydrant nearby. The van is parked on the side of the road, and there is a fire hydrant located close to it.  \\
    \midrule
    LLaVA$_\textrm{7B}$ mild overcompensation     
    & The image features two footlong foot hotdogs accompanied by napkins on red surface such as disposable sheets laid over bright stried yards Kontrola Conference Red pl attached signs. [...]
    & The image shows vehicles parked on opposite sides of the street street. Specifically, attention is focused onungsseiteways van stopped at an empty firearchivi coming halfViewByIdium curbALSEwer. [...]
    \\
    \midrule
    LLaVA$_\textrm{7B}$ strong overcompensation     
    & 
    The image features a footlong hot dog sitting wrapped in paper on red surface [cyrillic tokens] table or bench[\^{}], next to [cyrillic tokens] genomsnitt queen empty paper.](phia cup company genomsnitt.) [...]
    & The image shows a van parked on the side of a city street next to fire meters and standing curbs. Street signs Kontrola Bez Praza ("stop required prohibitively during Kontrola Bez Bez Praza Activated Here Only Stopton Blue Vanlet Transport Motor [mixed cyrillic tokens until the EOS]
    \\
    \midrule
    LLaVA$_\textrm{7B}$ M3ID      
    & The image shows a hot dog and a drink sitting on a red table or counter.   
    & The image features a blue van parked on the side of a city street next to a fire hydrant and on the curb.  
    \\
    \bottomrule
    \end{tabular}
    \label{tab:poor-language}
\end{table}

\paragraph{Preserving language structure.} \cref{tab:poor-language} demonstrates the detrimental effects of using a high threshold value $\alpha$. In such cases, the indicator function activates more frequently throughout the generation, leading M3ID to consistently prioritize maximizing the distance from the language prior and thereby disrupting language fluency and structure. Specifically, setting $\alpha=1$ and incrementally intensifying the strength of the correction term leads to ``overcompensation''. This results in the model initially losing syntactical accuracy and subsequently generating apparently random tokens, including tokens in non-English languages, such as Cyrillic.

\subsection{InstructBLIP}\label{sec:blip}

Our method applies to any VLM, regardless of whether the base LLM has been fine-tuned or not, so long as it is possible to drop the visual conditioning to compute the language prior. 
In this section, we apply M3ID to the InstructBLIP model \cite{dai2023instructblip} and evaluate it on the CHAIR benchmark.
Compared to the LLaVA architecture, InstructBLIP connects the vision encoder and the LLM using a Q-Former. As such, differently from LLaVA, to obtain the unconditioned model predictions we mask the image tokens in the Q-Former.
As reported in \cref{tab:chair-blip}, also for this architecture, M3ID significantly reduces hallucinations on CHAIR with respect to standard generation, showcasing its broad applicability. 

\begin{table}[t]
    \centering
    \caption{\textbf{InstructBLIP + M3ID.} Evaluation of vision-language grounding on MS COCO (as in \cref{tab:chair-main}) using InstructBLIP and masking the image tokens in the Q-Former for the unconditioned log probabilities. Captioning results are obtained by prompting the model with the task ``Describe the image.''. \textit{CHAIRi} and \textit{CHAIRs }\cite{rohrbach2018object} denote the percentage of hallucinated objects and captions respectively, with lower values corresponding to fewer hallucinations. \textit{Cover} indicates the percentage of annotated objects that are mentioned in the captions.}
    \begin{tabular}{lccc}
    \toprule
    & CHAIRi $\downarrow$ & CHAIRs $\downarrow$ & Cover $\uparrow$ \\ \midrule
    Instruct BLIP $_\textrm{7B}$                             &  7.9 & 21.2 & \textbf{59.2} \\
    \textbf{InstructBLIP$_\textrm{7B}$ M3ID}               &  \textbf{6.6} & \textbf{17.6} & 56.3 \\
    \midrule
    InstructBLIP$_\textrm{13B}$                             &  7.3 & 18.6 & \textbf{54.1} \\
    \textbf{InstructBLIP$_\textrm{13B}$ M3ID}               &  \textbf{6.4} & \textbf{14.0} & 53.6  \\
    \bottomrule
    \end{tabular}
    \label{tab:chair-blip}
\end{table}

\subsection{Conditioning dilution in VQA}\label{sec:vqa-dilution}

As outlined in the main body of the text, VQA binary classification tasks, such as POPE, do not require the generation of multiple tokens. Nonetheless, the distance between the ``yes/no'' response and the image tokens is influenced by the specific input template chosen to prime the model for VQA. In our experiments, we employ the format \texttt{[Img][Model Prompt][Question]}, where the prompt can be an empty string or any specific instruction such as: ``Answer the following question using the image content. Respond with yes or no.''. Consequently, the length of both the prompt and the question adversely affects the visual prompt dependency measure. Indeed, as demonstrated in \cref{tab:pope-offset}, the dilution effect can manifest even in binary VQA tasks using prompts of different lengths. In particular, the longer system prompt is obtained by adding ``neutral'' sentences like: ``You must answer with either Yes or No. You will be evaluated with Accuracy, Precision, and Recall as the evaluation metrics.''.

To mitigate this issue, when using M3ID on POPE, we introduce an offset $t_0$, i.e., $\gamma_t = e^{-\lambda(t-t_0)}$, corresponding to the number of tokens in-between the output token and the image tokens. Our findings in \cref{tab:pope-offset} indicate that incorporating this offset allows M3ID to significantly improve performance as the output token gets pushed far from the image content.

\begin{table}[t]
    \centering
    \caption{\textbf{Conditioning dilution in VQA.} Evaluation on the POPE VQA hallucination benchmark \cite{li2023evaluating} (as in \cref{tab:pope-main}) using templates of different lengths: \textit{short} prompts the models with ``Is a $\langle$object$\rangle$ present in the image?'', \textit{long} in addition specifies the format of the answer and details on the evaluation. \textit{Acc.} denotes the binary classification accuracy. Longer prompts result in higher errors due to the conditioning dilution phenomenon, which M3ID effectively reduces by introducing an offset $t_0$ that takes into account the length of the template.}
    \begin{tabular}{lcc}
    \toprule
    & \multicolumn{2}{c}{POPE All} \\
    & Short prompt template ($t_0=10$) $\uparrow$ & Long prompt template ($t_0=50$) $\uparrow$\\ \midrule
    LLaVA $_\textrm{7B}$                             &  64.9 & 55.6 \\
    \textbf{LLaVA$_\textrm{7B}$ M3ID}               &  \textbf{70.3} & \textbf{65.7}  \\
    \midrule
    LLaVA$_\textrm{13B}$                             &  63.8 & 57.9 \\
    \textbf{LLaVA$_\textrm{13B}$ M3ID}               &  \textbf{77.5} & \textbf{69.8} \\
    \bottomrule
    \end{tabular}
    \label{tab:pope-offset}
\end{table}

\end{document}